\documentclass[10pt,twocolumn,letterpaper]{article}
\usepackage[accsupp]{axessibility}  %

\usepackage[pagenumbers]{wacv} %

\usepackage[T1]{fontenc}
\usepackage{amssymb}
\usepackage{amsfonts}
\usepackage{amsmath}
\usepackage{amsxtra}
\usepackage{cancel}
\usepackage{dsfont}
\usepackage{graphicx}
\usepackage{tikz}
\usepackage{tikz-qtree}
\usetikzlibrary{shapes}
\usetikzlibrary{positioning}
\usetikzlibrary{trees}
\usepackage[most]{tcolorbox}
\usepackage{mathcomp}
\usepackage{mathtools}
\usepackage{multirow}
\usepackage{verbatim}
\usepackage{polynom}
\usepackage{textcomp}
\usepackage{float}
\usepackage{xcolor}
\usepackage{pdflscape}
\usepackage{csquotes}
\usepackage{afterpage}
\usepackage{makecell}
\usepackage{listings}
\usepackage{url}
\usepackage{enumitem}
\usepackage{minibox}
\usepackage{shuffle}
\usepackage{pifont}
\usepackage{subcaption}
\usepackage{xspace}
\usepackage{siunitx}
\usepackage{booktabs}
\usepackage[protrusion=true,expansion=true]{microtype}
\usepackage{colortbl}
\usepackage{footmisc}
\usepackage[export]{adjustbox}
\usepackage{placeins}

\DeclareMathSymbol{\mlq}{\mathord}{operators}{``}
\DeclareMathSymbol{\mrq}{\mathord}{operators}{`'}

\newcommand{\R}{\mathbb R}

\renewcommand{\phi}{\varphi}

\newcommand{\bQ}{\mathbf{Q}}

\definecolor{darkgreen}{HTML}{009B55}

\newcommand{\code}[1]{\texttt{#1}}
\newcommand{\cmark}{\ding{51}}%
\newcommand{\xmark}{\ding{55}}%
\newcommand{\rowhighlight}{\rowcolor{gray!30}}

\newtcolorbox{takeaway}[1][Takeaway]{
	colback=blue!4!white,      %
	colframe=blue!50!black,    %
	boxrule=0.6pt,
	arc=2pt,
	left=6pt,right=6pt,top=2pt,bottom=2pt,
	title=#1,
	fonttitle=\bfseries\small,
	coltitle=white,
	colbacktitle=blue!50!black
}

\definecolor{wacvblue}{rgb}{0.21,0.49,0.74}
\usepackage[pagebackref,breaklinks,colorlinks,allcolors=wacvblue]{hyperref}

\def\wacvPaperID{67} %
\def\confName{WACV}
\def\confYear{2027}

\newcommand{\name}{LUMA\xspace}
\newcommand{\longname}{Lightweight Universal Mask Adapter\xspace}

\title{\name: Benchmarking Segmentation via a \longname}

\author{Tobias Christian Nauen${}^{1, 2}$, Anosh Billimoria${}^1$, Federico Raue${}^2$,\\ Stanislav Frolov${}^2$, Brian B. Moser${}^2$, Andreas Dengel${}^{1, 2}$\\
${}^1$ RPTU University Kaiserslautern-Landau, Kaiserslautern, Germany \\
${}^2$ German Research Center for Artificial Intelligence (DFKI), Kaiserslautern, Germany \\
{\tt\small first\_second.last@dfki.de / first.last@dfki.de}
}

\begin{document}
\maketitle
\begin{abstract}
	Comparing transformer backbones for image segmentation is confounded: each is paired with a different decoder, recipe, and pretraining, so reported differences rarely reflect the backbone itself.
	We introduce the \longname (\name), a lightweight, backbone-agnostic mask-transformer head that treats any backbone as a black-box feature extractor, letting a set of queries read from its features through cheap cross-attention.
	\name matches the accuracy of EoMT, the state-of-the-art efficient ViT-segmenter, at cost parity, while attaching unchanged to hierarchical, convolutional, and mixture-of-experts backbones alike.
	Holding this head fixed, we benchmark 20 backbones, 11 pretraining schemes and a range of resolutions on ADE20K, COCO, and Cityscapes under one modern recipe.
	We find that ``efficient'' token mixers fail to deliver efficiency even at the high resolutions that motivate them, with plain ViT holding the throughput Pareto front at every resolution.
	Additionally, the pretraining objective, not the architecture, the lever the field has tuned hardest, governs segmentation quality.
	\\
	Code: \url{https://github.com/tobna/LUMA}
\end{abstract}

\section{Introduction}
\label{sec:intro}

Modern image segmentation is built on the mask transformer framework~\cite{Cheng2021,Cheng2022}, where a set of learnable queries is matched against the features of a pretrained backbone and decoded into a mask and class per segment.
Many efforts have since optimized the mask-based segmentation-decoder~\cite{Li2023g, Jain2023, Yu2022a, Li2022c}.
Recently, EoMT~\cite{Kerssies2025} showed that this task-specific decoder part of the pipeline is largely unnecessary for \emph{plain ViT}~\cite{Dosovitskiy2021}.
With a strong pretraining, the ViT backbone's own self-attention can process the queries directly.
Thus, in modern pipelines, most of the compute sits in the ViT backbone, for which a decade of work has now gone into optimizing: efficient and linear attention~\cite{Wang2020,Ma2024,Xiong2021}, hierarchical and windowed token mixers~\cite{Liu2021}, convolutional hybrids~\cite{Wu2021,Zhang2021,Cai2023,Li2022b}, and state-space models~\cite{Liu2024c}; almost all of these justified by efficiency at high resolution.
This invites the question: Is the common backbone \emph{architecture} of ViT itself necessary, or do pretraining and scale dominate dense transfer, and we can thus reap the efficiency gains of efficient backbones at high resolutions?

Answering this requires comparing backbones \emph{on equal footing}:
\emph{One} segmentation head, that attaches to any backbone, adds only little compute, and reaches state-of-the-art accuracy. Trained with \emph{one} training recipe, with only the backbone varying, so that measured differences reflect the architecture rather than the surrounding apparatus.
Existing comparisons do not meet this bar, as each backbone is typically paired with its own decoder, recipe, pretraining, and resolution~\cite{Li2024g, Goldblum2023} and the heavy task-specific decoders that dominate the leaderboard confound the backbone with the capacity of the head itself~\cite{Cheng2022, Chen2023c}.
EoMT, sporting only a very lightweight mask-head, is the natural fair instrument, but its efficiency comes from concatenating the queries \emph{into} the token sequence and processing them with the backbone's own attention, which presumes global attention, reach-in access to the block internals, and a constant token width.
It is therefore a \emph{plain-ViT} method by construction, and cannot be attached to the very backbones we wish to study.

We close this gap with the \longname (\name), a lightweight, efficient segmentation head that treats the backbone as a black-box feature extractor.
Rather than inserting queries into the token stream, \name keeps them in a separate side stream and lets them \emph{read} from each tapped block's patch features through a lightweight cross-attention block, reusing the backbone's own feed-forward layers, so the queries inherit any width schedule.
The backbone runs exactly as it would on its own, so the identical head attaches unchanged to isotropic, hierarchical, windowed, convolutional, and mixture-of-experts backbones alike.
This decoupling also yields a result complementary to EoMT's:
Efficient, high-accuracy segmentation can be achieved with a minimal external read of the backbone's features, which is precisely what frees us from the global-attention and shared-width assumptions.

With \name held fixed as a fair measuring stick, we conduct a controlled study of 20 vision transformer backbones, 11 pretraining schemes and a range of input resolutions on ADE20K~\cite{Zhou2019}, Cityscapes~\cite{Cordts2016}, and COCO~\cite{Lin2014_Microsoft_COCO_Comm}, in semantic and panoptic segmentation; all under a single modern recipe.
This lets us disentangle, for the first time under one modern protocol, how segmentation quality is governed by the backbone architecture, pretraining objective, and by resolution.

Our dominant finding concerns the very property these architectures are designed for.
``Efficient'' token mixers exist to escape the $\mathcal{O}(T^2)$ token scaling of attention, a cost that is negligible at the short sequences of $224$\,px classification but should dominate at the $512$--$1024$\,px resolutions segmentation tasks use.
Nonetheless, even without FlashAttention~\cite{Dao2022,Dao2024}, the plain ViT holds the throughput Pareto front, with efficient variants competitive only in memory.
On the accuracy side, differences from architecture largely vanish once every backbone hosts the same head, with a few mixers (Linformer~\cite{Wang2020}, Synthesizer~\cite{Tay2021}, MLP-Mixer~\cite{Tolstikhin2021}) failing to generalize to high resolution; a failure classification benchmarks never expose.
In line with this, ImageNet top-1 accuracy is an unreliable predictor of segmentation quality across backbones.
Orthogonally, the pretraining objective is the important part: dense objectives transfer robustly while supervised pretraining is a categorical outlier despite competitive classification accuracy.

In summary: \textbf{(1)} we introduce \name, a lightweight, backbone-agnostic mask transformer head that matches EoMT's accuracy and efficiency while attaching unchanged to \emph{any} token mixer.
\textbf{(2)} Using \name as a fixed instrument, we present the broadest fair benchmark of transformer backbones for segmentation to date, spanning 20 architectures, 11 pretraining schemes, multiple resolutions, and semantic and panoptic tasks under one recipe.
\textbf{(3)} Our analysis disentangles architecture, pretraining, and resolution, and shows that token-mixer design is a weak lever for dense prediction while the pretraining objective and plain scale are the strong ones, and that the plain ViT remains the backbone of choice.

\section{Related Work}

\textbf{Segmentation with Transformers.}
Segmentation was initially dominated by CNNs, whose feature maps naturally yield dense, pixel-level predictions~\cite{Long2015, Ronneberger2015, Chen2018a}.
After the introduction of ViT~\cite{Dosovitskiy2021}, custom variants were developed specifically for segmentation~\cite{Xie2021}.
A key development was the MaskFormer family~\cite{Cheng2021, Cheng2022}, which reframed segmentation as predicting a set of masks with associated class labels rather than classifying each pixel independently, and now underpins most state-of-the-art segmenters~\cite{Li2023g, Jain2023, Yu2022a}.
To utilize plain ViTs for segmentation, heavy pixel decoders were appended after the backbone~\cite{Strudel2021, Cheng2021, Li2022c, Li2023g} and adapters reaching into its intermediate stages were introduced~\cite{Chen2023c, Jain2023, Cheng2022}.
Recently, EoMT~\cite{Kerssies2025} showed that this machinery is largely unnecessary, repurposing the backbone's own self-attention as the adapter to match MaskFormer-like region queries to image patches with only a lightweight mask module on top.
Our \name{} continues this trend toward minimal segmenters, generalizing the EoMT setup beyond plain self-attention backbones to arbitrary token mixers, while keeping the head lightweight via only a few cross-attention layers.

\textbf{Vision Transformer Backbones.}
The $\mathcal{O}(T^2)$ scaling of self-attention in the number of tokens $T$ makes the plain ViT~\cite{Dosovitskiy2021} costly at the high resolutions segmentation demands, motivating a wide range of architectural modifications.
Following the taxonomy of efficient transformer variants by \citet{Nauen2025}, we benchmark low-rank attention (Linformer~\cite{Wang2020}, Nyströmformer~\cite{Xiong2021}), sparse attention (Swin~\cite{Liu2021}, Wave-ViT~\cite{Yao2022}), fixed attention (Synthesizer~\cite{Tay2021}), kernel attention (Poly-SA~\cite{Babiloni2023}, Hydra~\cite{Bolya2022}), and hybrid attention (EfficientViT~\cite{Cai2023}, Next-ViT~\cite{Li2022b}, CvT~\cite{Wu2021}, ResT~\cite{Zhang2021}, CoaT~\cite{Xu2021}); non-attention token mixers that replace attention entirely (MLP-Mixer~\cite{Tolstikhin2021}, EfficientMod~\cite{Ma2024}, FocalNet~\cite{Yang2022a}); and models that modify the feed-forward block (Switch~\cite{Fedus2022}, HiViT~\cite{Zhang2023c}).
For a broader overview of efficient transformers, we refer to~\cite{Nauen2025, Liu2023b, Patro2023}.

\textbf{Pretraining Strategies.}
Backbones are typically initialized from weights pretrained on a large dataset, and this choice of pretraining objective strongly affects downstream segmentation quality~\cite{Kerssies2024}.
Traditionally, computer vision pretraining relied on large-scale supervised image classification~\cite{Deng2009, He2016, Yosinski2014}, for which the dominant transformer recipes are DeiT~\cite{Touvron2021b} and DeiT~III~\cite{Touvron2022}.
More recent objectives move beyond label supervision: image--text contrastive learning (CLIP~\cite{Radford2021}, SigLIP~\cite{Zhai2023}, AIMv2~\cite{Fini2024}), masked image modeling (MAE~\cite{He2022}, BEiT-3~\cite{Wang2023b}, EVA-02~\cite{Fang2024}), the self-distillation family of DINO~\cite{Caron2021, Oquab2024, Simeoni2025}, self-supervised segmentation pretraining SAM~\cite{Kirillov2023} and recently the joint-embedding predictive I-JEPA~\cite{Assran2023}.
We benchmark representatives of all families.

\textbf{Benchmarking Segmentation.}
Several prior works benchmark components of the segmentation pipeline in isolation.
\citet{Ranftl2021, Jeeveswaran2022} compare CNN and transformer backbones, but predate modern mask-based segmenters and large-scale pretraining.
\citet{Li2024g} benchmark segmentation decoders while holding the backbone fixed, and \citet{Kerssies2024} benchmark pretraining objectives using a lightweight linear-probe and heavy Mask2Former setup.
BoB~\cite{Goldblum2023} benchmarks backbones for multiple tasks, including segmentation, but they conflate architecture and pretraining differences, as ``different pretraining algorithms were trained on different datasets and architectures''~\cite{Goldblum2023}.
\citet{Bensaid2025} target the few-shot setting, and \citet{Agnihotri2025} evaluate segmentation robustness.
In contrast, we benchmark 20 backbone architectures, using the same pretraining setup, and 11 pretraining schemes under a single modern recipe, plugging each into \name, a lightweight, backbone-agnostic, MaskFormer-based segmentation head.
This lets us disentangle architecture, pretraining, and resolution under a fair, modern segmentation protocol.

\section{Method}
\label{sec:method}

\begin{figure}
	\centering
	\includegraphics[width=0.95\columnwidth]{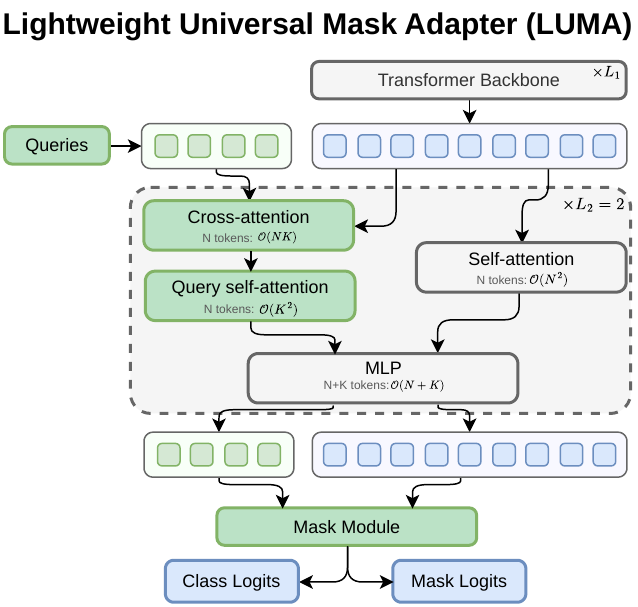}
	\caption{\textbf{\name architecture.}
		The backbone (gray) runs unmodified on the $N$ patch tokens. A side-stream of $K$ queries reads from each tapped block's features via cross-attention.} \label{fig:luma-method}
\end{figure}

Our goal is to benchmark a broad range of transformer backbones for image segmentation \emph{on equal footing}.
This requires a segmentation head that (i) attaches to any backbone without architecture-specific surgery, (ii) adds as little overhead as possible so that measured differences reflect the \emph{backbone} rather than the head, and (iii) reaches the accuracy of state-of-the-art mask transformers.
No existing design satisfies all three at once: heavy task-specific decoders inflate and confound the head, while the recent EoMT~\cite{Kerssies2025} is lightweight but tightly coupled to the plain ViT and cannot be applied to most backbones.
We therefore introduce the \longname (\name), a minimal head that treats the backbone as a black-box feature extractor and lets a small set of queries \emph{read} from its features through lightweight cross-attention.
\name matches EoMT in accuracy, and efficiency, but applies to \emph{any} backbone family.

\subsection{Preliminaries}
\label{sec:method:prelim}

A transformer backbone maps an image $\mathbf{I}\in\mathbb{R}^{3\times H\times W}$ to a sequence of feature tokens through $L$ blocks.
Backbones differ widely in how they do this: plain ViTs~\cite{Dosovitskiy2021} keep a constant token count and embedding dimension and use \emph{global} self-attention; hierarchical backbones such as Swin~\cite{Liu2021} downsample across stages, grow the embedding dimension, and use \emph{windowed} or local attention; and recent state-space and convolutional hybrids replace attention entirely~\cite{Li2022b, Yang2022a, Liu2024c}.
A backbone-agnostic head must cope with all of these without assuming any particular attention mechanism, token layout, or dimension schedule.

Modern segmentation follows the mask transformer framework~\cite{Cheng2021}: a set of $K$ learnable queries $\bQ \in \R^{K \times D}$ each learn to represent a single segment (a \emph{thing} instance or \emph{stuff} class).
The queries are refined against image features and then decoded into a class label and a binary mask per query.
Class logits $\mathbf{c}_i \in \mathbb{R}^{C}$ are obtained with a linear layer, and mask logits are obtained by passing each query through an MLP
and taking the dot product with high-resolution image features.
During training, one assigns each ground-truth segment to a unique query and supervises both predictions.
State-of-the-art models such as Mask2Former~\cite{Cheng2022} refine queries using a heavy stack of multi-scale deformable cross-attention and self-attention.

\subsection{EoMT and the backbone-coupling problem}
\label{sec:method:eomt}

EoMT~\cite{Kerssies2025} removes the decoder altogether.
It concatenates the $K$ queries \emph{into} the token sequence after the first $L_1 = L - L_2$ blocks, and lets the final $L_2$  out of $L$ ViT blocks process queries and patch tokens \emph{jointly} through the ViT's own self-attention.
During training, intermediate masks are predicted before each block and used to mask the query-to-patch attention, mimicking the masked cross-attention of Mask2Former; a mask-annealing schedule then phases this masking out so that inference needs no masking anymore.
The result is elegant and efficient: the head is little more than a handful of extra tokens and a small mask module.

This efficiency, however, is purchased with implicit assumptions about the backbone:
(1)~\textbf{Global attention.} Because the queries are appended to the token sequence and must attend to \emph{all} patches, the backbone's attention must be global; windowed or neighborhood attention would confine each query to a single window.
(2)~\textbf{Accessible attention internals.} To inject the attention mask, EoMT re-implements the block's attention.
This ties it to the plain ViT block's internals and needs to be re-implemented for any customized attention block.
In short, EoMT is a \emph{plain-ViT-only} method by construction.
It is thus unsuitable as a benchmarking instrument, since most backbones we want to compare cannot host it.

\subsection{\name: a \longname}
\label{sec:method:luma}

The key idea behind \name is to make the queries backbone-agnostic, by decoupling them from the backbone's token-mixing mechanism.
Instead of inserting queries into the token stream and relying on the backbone's attention to mix them with image features, \name keeps the queries in a separate side-stream and gives them a cheap, explicit way to read from the backbone's patch features.
The backbone is run exactly as it would be on its own, block by block, and is never modified, re-implemented, or even aware of the queries.
An overview is shown in \Cref{fig:luma-method}.

\textbf{Query side-stream.}
Let $\mathbf{X}^{i}$ denote the patch features produced by backbone block $i$, and let $\mathbf{Q}^{i}$ denote the query states, initialized from a learnable embedding $\mathbf{Q}^{L_1}$.
For each such block $i > L_1$, \name updates the queries in two steps.
First, a single lightweight cross-attention layer lets the queries read from the \emph{current} patch features, which supply the keys and values.
Second, the queries are passed through a self-attention layer and block $i$'s \emph{own} feed-forward sub-layer, reusing the backbone's MLP with its existing weights:
\begin{align}
	\begin{split}
		\bQ_\text{xa}^{i} & = \bQ^{i} + \texttt{LS}(\texttt{XAttn}(\bQ^{i},\,\mathbf{X}^{i})),                  \\
		\bQ_\text{sa}^i   & = \bQ_\text{xa}^i + \texttt{LS}(\texttt{SAttn}(\bQ_\text{xa}^i))                    \\
		\mathbf{Q}^{i+1}  & = \bQ_\text{sa}^{i} + \texttt{MLP}_i\!\big(\texttt{Norm}_i(\bQ_\text{sa}^{i})\big),
	\end{split}
	\label{eq:luma-update}
\end{align}
where $\texttt{XAttn}$ is cross-attention, $\texttt{SAttn}$ is self-attention, and $\texttt{LS}$ is a per-channel LayerScale~\cite{Touvron2021a}.
Together these form the \emph{only} new parameters \name introduces at block $i$.
The normalization $\texttt{Norm}_i$ and feed-forward layer $\texttt{MLP}_i$ are block $i$'s own, applied to the queries with exactly the weights used for the patch tokens.
Backbones without a per-block MLP, such as ResNet, instead get a separate MLP for the queries, which performs similarly (see ablation in \Cref{apdx:luma-ablation}).
For \emph{panoptic segmentation} training we use EoMT's mask-annealing schedule, but apply the mask to the cross-attention read instead of reaching into the backbone's self-attention.
The patch stream is left untouched, $\mathbf{X}^{i+1} = \texttt{Block}_i(\mathbf{X}^{i})$, so the queries influence the backbone only by \emph{reading} from it through \Cref{eq:luma-update}.

Reusing the backbone's feed-forward sub-layer also dictates how \name handles a changing token dimension.
Because the queries are processed by block $i$'s MLP and cross-attend to its patch features, $\mathbf{Q}^{i}$ and $\mathbf{X}^{i}$ must share the same width.
For plain ViTs this holds trivially, but hierarchical backbones change their internal dimension between stages through an explicit mechanism (e.g.\ the projection within patch merging~\cite{Liu2021}).
Rather than add a projection of our own, which would be redundant and a source of architecture-specific tuning, \name routes the queries through the \emph{same} dimension-changing operation the backbone applies to its tokens.
The queries thus inherit the backbone's width schedule for free and stay dimension-aligned with the patch features, regardless of how their width changes.

\textbf{Mask prediction module.}
For mask and class prediction we follow Mask2Former~\cite{Cheng2022} and EoMT: a linear layer maps each final query to class logits, and a three-layer MLP maps it to a mask embedding.
Mask logits are the dot product between mask embeddings and a single-scale image feature map upscaled by a small \texttt{ScaleBlock} stack~\cite{Li2022c}.
To obtain this feature map from an arbitrary backbone, \name reconstructs the spatial grid from the backbone's output according to its specifications.

\textbf{Backbone interaction.}
\name interacts with a backbone only through a small, generic interface: it needs the per-block features, together with a position map to interpret which features correspond to which image regions.
It never assumes global attention, never reads the block's attention internals, and doesn't require a constant width.
As a result the same head attaches unchanged to plain ViTs, hierarchical and windowed backbones, convolutional hybrids, and even to mixture-of-experts backbones or CNNs.
This makes \name a fair benchmarking instrument: the head is lightweight and held fixed while only the backbone varies.

It would be tempting to read \name as ``re-adding the decoder that EoMT removed,'' but it is neither that nor a trivial rearrangement.
First, \name is not a multi-scale pixel decoder plus transformer decoder; it is a \emph{single} cross-attention read per tapped block on \emph{single-scale} features, with no deformable attention and no feature fusion.
Second, achieving generality while preserving EoMT's accuracy and efficiency is not obvious a priori.
EoMT's central claim was that the ViT's own self-attention is \emph{sufficient} to integrate queries, with no external module needed.
Now, \name shows that a minimal \emph{external} read recovers the same accuracy and efficiency, which is what makes it possible to drop the global-attention and shared-width assumptions.

\textbf{Compute cost.}
\name is as lightweight as EoMT, which concatenates the $K$ queries into the token sequence, so each of its final blocks runs self-attention over $N+K$ tokens, at cost $\mathcal{O}\big((N+K)^2\big)$.
\name instead leaves the backbone's self-attention to operate on the $N$ patch tokens alone and adds only cross-attention and query self-attention, $\mathcal{O}(NK + K^2)$, for a per-block cost of $\mathcal{O}(N^2 + NK + K^2)$.
On top of this, the dedicated read lets \name use \emph{fewer} query-processing blocks: we use $L_2 = 2$, against $L_2 = 4$ for EoMT.
Aside from the attention, in \name and EoMT the queries pass through the last blocks' feed-forward layers, adding $K \ll N$ tokens to a pointwise operation.

\section{Experiments}
We first compare \name to current segmentation pipelines in \Cref{sec:luma-segmentation}.
Then we benchmark different configurations for efficient segmentation using the Pareto front, the boundary of the landscape of efficient models, as our framework for analysis.
We benchmark 20 transformer backbones, all trained under the same state-of-the-art DeiT~III~\cite{Touvron2022} recipe on ImageNet-21k~\cite{Ridnik2021}, in \Cref{sec:backbone}, different pretraining setups in \Cref{sec:pretraining} and Pareto-optimal compute scaling in \Cref{sec:scaling}.
For the full hyperparameters and ablations of \name, see \Cref{apdx:setup,apdx:luma-ablation}.

\subsection{Segmentation with \name}
\label{sec:luma-segmentation}
\begin{table*}[ht!]
	\caption{\textbf{\name for semantic segmentation.} With backbone and pretraining from DINOv2, \name matches strong mask-transformer baselines while being lightweight and \emph{backbone agnostic}. ViT-Adapter + Mask2Former, EoMT, and \name use windowed inference, dividing each image into multiple crops, and the FLOPs and FPS results account for this. $^{\dagger}$On ADE20K, these models resize the shortest side of images to the indicated scale during inference, while preserving aspect ratio. $^{\ddagger}$Our re-implementation. The rest is taken directly from~\citet{Kerssies2025} whose training and evaluation protocol we follow.}
	\label{tab:luma-semantic-seg}
	\centering
	\setlength{\tabcolsep}{4pt}
	\resizebox{\textwidth}{!}{
		\begin{tabular}{lllc c S[table-format=4.0] S[table-format=3.0] S[table-format=2.1] c S[table-format=4.0] S[table-format=3.0] S[table-format=2.1]}
			\toprule
			\multirow{2.5}{*}{Method}                & \multirow{2.5}{*}{Backbone}    & \multirow{2.5}{*}{Pre-training} & \multirow{2.5}{*}{Params} & \multicolumn{4}{c}{Cityscapes \textit{val} \cite{Cordts2016}} & \multicolumn{4}{c}{ADE20K \textit{val} \cite{Zhou2019}}                                                   \\
			\cmidrule(lr){5-8} \cmidrule(lr){9-12}
			                                         &                                &                                 &                           & Input size                                                    & GFLOPs                                                  & FPS  & mIoU & Input size & GFLOPs & FPS  & mIoU \\
			\midrule
			Mask2Former$^{\dagger}$ \cite{Cheng2022} & Swin-L \cite{Liu2021}          & IN21K                           & 216M                      & $1024 \times 2048$                                            & {--}                                                    & 14   & 83.3 & $640^2$    & {--}   & 33   & 56.1 \\
			MaskDINO$^{\dagger}$ \cite{Li2023g}      & Swin-L \cite{Liu2021}          & IN21K                           & 223M                      & {--}                                                          & {--}                                                    & {--} & {--} & $640^2$    & {--}   & {--} & 56.6 \\
			OneFormer$^{\dagger}$ \cite{Jain2023}    & ConvNext-XL \cite{Liu2022}     & IN21K                           & 373M                      & $1024 \times 2048$                                            & 775                                                     & 7    & 83.6 & $640^2$    & 607    & 21   & 57.4 \\
			OneFormer$^{\dagger}$ \cite{Jain2023}    & DiNAT-L \cite{Hassani2023}     & IN21K                           & 223M                      & $1024 \times 2048$                                            & 450                                                     & 14   & 83.1 & $896^2$    & 678    & 19   & 58.1 \\
			kMaX-DeepLab \cite{Yu2022a}              & ConvNext-L \cite{Liu2022}      & IN21K                           & 232M                      & $1025 \times 2049$                                            & 1673                                                    & {--} & 83.5 & {--}       & {--}   & {--} & {--} \\
			Mask2Former \cite{Cheng2022}             & ViT-L \cite{Dosovitskiy2021}   & DINOv2 + DA \cite{Yang2024a}    & {--}                      & $896 \times 1792$                                             & {--}                                                    & {--} & 84.8 & $896^2$    & {--}   & {--} & 59.4 \\
			\midrule
			Mask2Former \cite{Cheng2022}             & ViT-Adapter-L \cite{Chen2023c} & DINOv2                          & 351M                      & $1024^2$                                                      & 5200                                                    & 7    & 84.5 & $512^2$    & 910    & 21   & 58.9 \\
			EoMT${}^\ddagger$ \cite{Kerssies2025}    & ViT-L \cite{Dosovitskiy2021}   & DINOv2                          & 319M                      & $1024^2$                                                      & 4350                                                    & 36   & 84.2 & $512^2$    & 789    & 150  & 58.4 \\
			\midrule
			\textbf{\name} (Ours)                    & ViT-L \cite{Dosovitskiy2021}   & DINOv2                          & 336M                      & $1024^2$                                                      & 4371                                                    & 37   & 84.6 & $512^2$    & 790    & 143  & 58.6 \\
			\bottomrule
		\end{tabular}}
\end{table*}

\begin{table*}[t]
	\centering
	\caption{%
		\textbf{\name for panoptic segmentation.}
		With the backbone and pretraining fixed to DINOv2, \name again performs on-par with EoMT.
		\textsuperscript{\dag}During inference, these models resize the shortest side of images to the indicated scale, while preserving the aspect ratio.
		\textsuperscript{c}Models for these ADE20K results are pre-trained for COCO panoptic segmentation.%
		\textsuperscript{\ddag}Our re-implementation. The rest is taken from~\cite{Kerssies2025}.
	}
	\label{tab:luma-panoptic-seg}
	\setlength{\tabcolsep}{4pt}
	\resizebox{.95\textwidth}{!}{
		\begin{tabular}{
				l l l c
				c S[table-format=4.0] S[table-format=3.0] S[table-format=2.1]
				c S[table-format=4.0] S[table-format=3.0] S[table-format=2.1, table-space-text-post={\textsuperscript{c}}]
			}
			\toprule
			\multirow{2.5}{*}{Method}                          & \multirow{2.5}{*}{Backbone}    & \multirow{2.5}{*}{Pre-training} & \multirow{2.5}{*}{Params} & \multicolumn{4}{c}{COCO \emph{val2017}~\cite{Lin2014_Microsoft_COCO_Comm}} & \multicolumn{4}{c}{ADE20K \emph{val}~\cite{Zhou2019}}                                                                            \\
			\cmidrule(lr){5-8} \cmidrule(lr){9-12}
			                                                   &                                &                                 &                           & {Input size}                                                               & {GFLOPs}                                              & {FPS} & {PQ} & {Input size} & {GFLOPs} & {FPS} & {PQ}                    \\
			\midrule
			Mask2Former\textsuperscript{\dag}~\cite{Cheng2022} & Swin-L~\cite{Liu2021}          & IN21K                           & 216M                      & $800^2$                                                                    & 868                                                   & 24    & 57.8 & $640^2$      & {--}     & 33    & 48.1                    \\
			kMaX-DeepLab~\cite{Yu2022a}                        & ConvNeXt-L~\cite{Liu2022}      & IN21K                           & 232M                      & $1281^2$                                                                   & {--}                                                  & {--}  & 58.0 & $1281^2$     & 1302     & {--}  & 50.9                    \\
			OneFormer\textsuperscript{\dag}~\cite{Jain2023}    & DiNAT-L~\cite{Hassani2023}     & IN21K                           & 223M                      & $800^2$                                                                    & 736                                                   & 20    & 58.0 & $1280^2$     & 1369     & 10    & 51.5                    \\
			OneFormer\textsuperscript{\dag}~\cite{Jain2023}    & DiNAT-L~\cite{Hassani2023}     & IN21K                           & 223M                      & {--}                                                                       & {--}                                                  & {--}  & {--} & $1280^2$     & 1369     & 10    & 53.5\textsuperscript{c} \\
			MaskDINO\textsuperscript{\dag}~\cite{Li2023g}      & Swin-L~\cite{Liu2021}          & IN21K                           & 223M                      & $800^2$                                                                    & 1326                                                  & 14    & 58.3 & {--}         & {--}     & {--}  & {--}                    \\
			\midrule
			Mask2Former~\cite{Cheng2022}                       & ViT-Adapter-L~\cite{Chen2023c} & DINOv2                          & 349M                      & $640^2$                                                                    & 830                                                   & 29    & 57.1 & $640^2$      & 830      & 29    & 51.8\textsuperscript{c} \\
			Mask2Former~\cite{Cheng2022}                       & ViT-Adapter-L~\cite{Chen2023c} & DINOv2                          & 354M                      & $1280^2$                                                                   & 4817                                                  & 10    & 59.7 & $1280^2$     & 4817     & 10    & 53.0\textsuperscript{c} \\
			EoMT\textsuperscript{\ddag}~\cite{Kerssies2025}    & ViT-L~\cite{Dosovitskiy2021}   & DINOv2                          & 317M                      & $640^2$                                                                    & 750                                                   & 128   & 56.0 & $640^2$      & 750      & 128   & 50.6\textsuperscript{c} \\
			EoMT\textsuperscript{\ddag}~\cite{Kerssies2025}    & ViT-L~\cite{Dosovitskiy2021}   & DINOv2                          & 322M                      & $1280^2$                                                                   & 4146                                                  & 42    & 58.3 & $1280^2$     & 4146     & 42    & 51.7\textsuperscript{c} \\
			\midrule
			\textbf{\name} (Ours)                              & ViT-L~\cite{Dosovitskiy2021}   & DINOv2                          & 333M                      & $640^2$                                                                    & 754                                                   & 105   & 55.8 & $640^2$      & 754      & 105   & 50.4                    \\
			\textbf{\name} (Ours)                              & ViT-L~\cite{Dosovitskiy2021}   & DINOv2                          & 338M                      & $1280^2$                                                                   & 4469                                                  & 32    & 58.3 & $1280^2$     & 4469     & 32    & 51.7                    \\
			\bottomrule
		\end{tabular}}
\end{table*}

To ensure fair and meaningful benchmark results, we first compare \name against representative, modern mask-transformer segmenters on Cityscapes~\cite{Cordts2016} and ADE20K~\cite{Zhou2019} for semantic segmentation in \Cref{tab:luma-semantic-seg} and on COCO~\cite{Lin2014_Microsoft_COCO_Comm} and ADE20K~\cite{Zhou2019} for panoptic segmentation in \Cref{tab:luma-panoptic-seg}.
To make this a test of the \emph{segmentation mechanism} rather than of the backbone or pre-training, we fix the backbone to the canonical ViT-L/DINOv2~\cite{Oquab2024} configuration shared by EoMT and ViT-Adapter\,+\,Mask2Former.

The three DINOv2 plain-ViT models cluster tightly on both benchmarks ($84.2$--$84.6$ / $58.4$--$58.9$ for semantic and $58.3$--$59.7$ / $51.7$--$53.0$ for panoptic segmentation) and match or exceed the IN21K systems despite far simpler heads.
Among the DINOv2 group, \name matches EoMT as the lightest:
On ADE20K it uses $1.2\times$ fewer GFLOPs than ViT-Adapter\,+\,M2F and runs $\approx 7\times$ faster.
\name reaches $84.6$\,mIoU on Cityscapes and $58.6$\,mIoU on ADE20K, against EoMT's $84.2$ and $58.4$, and $58.3$\,PQ on COCO and $51.7$\,PQ on ADE20K against EoMT's $58.3$ and $51.7$.
This performance also comes at similar cost in terms of GFLOPs and throughput.
The result is expected from the design.
EoMT's in-sequence formulation is already optimized \emph{for ViT}, and \name matches it.

The parity above matters precisely because of the flexibility \name offers over the baselines.
EoMT obtains its low overhead by concatenating queries into the patch-token sequence and processing them with the backbone's own self-attention---a formulation defined only for plain ViT and therefore inapplicable to the hierarchical, mixture-of-experts, linear-attention, and non-attention backbones we study in our benchmark.
\name instead confines query--image interaction to external cross-attention blocks, decoupling the head from the mixing method.
This decoupling is what lets the \emph{identical} head attach to every backbone in our benchmark.

\subsection{Which transformer to use?}
\label{sec:backbone}
\begin{figure}
	\centering
	\includegraphics[width=0.95\columnwidth]{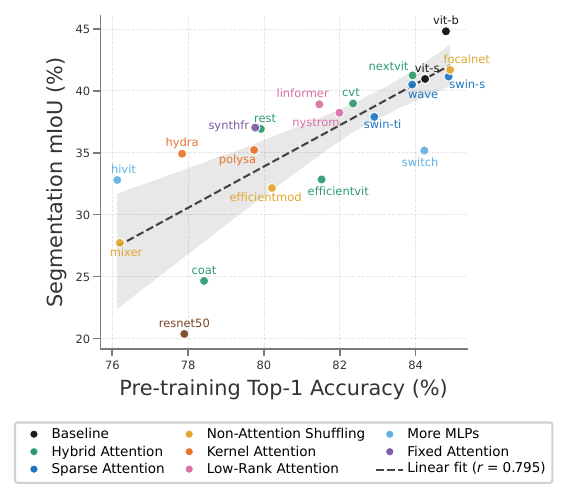}
	\caption{\textbf{ImageNet top-1 is an unreliable predictor of segmentation quality.} ADE20K mIoU at $224$\,px vs ImageNet-1k top-1 ($r \approx 0.80$). Most models cluster near the trend; models below the trend often incorporate convolutions.}\label{fig:backbones-pretrain-vs-iou}
\end{figure}

\begin{figure}
	\centering
	\includegraphics[width=\columnwidth]{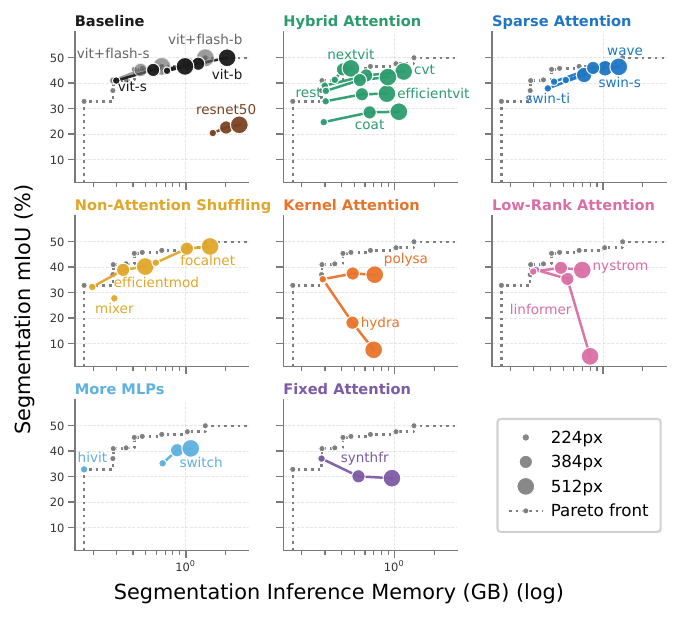}
	\caption{\textbf{Inference Memory vs.\ ADE20K mIoU across backbone taxonomy.} Panels group families from~\cite{Nauen2025}; markers indicate image resolution. Plain ViT holds the Pareto front; among efficient backbones mostly sparse attention (Swin-S, WaveViT) is close. Several mixers degrade as resolution rises.}\label{fig:pareto-memory-ade}
\end{figure}

\begin{figure}
	\centering
	\includegraphics[width=\columnwidth]{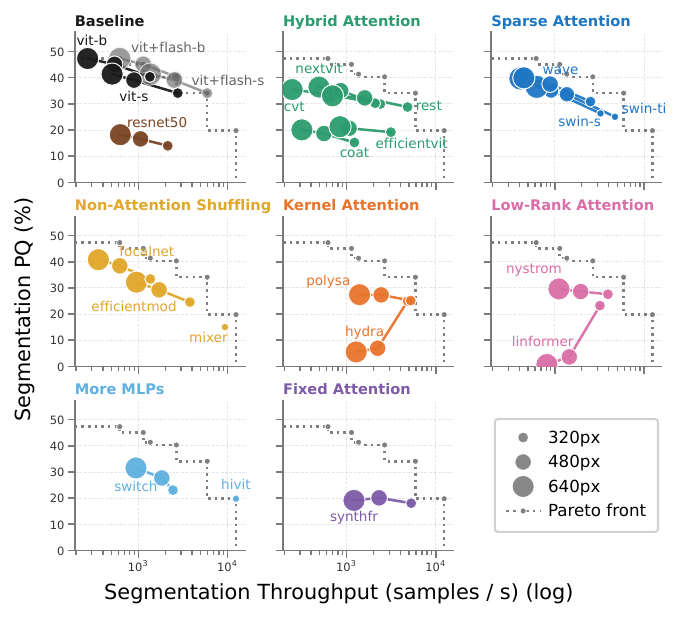}
	\caption{\textbf{Throughput vs.\ COCO PQ across backbone taxonomy.} Plain ViT is again Pareto-optimal, but on the slower end when not using FlashAttention. Among efficient backbones, sparse attention is close to the Pareto-front. Hybrid attention does not scale as well with resolution as ViT, sparse, and non-attention methods do.}\label{fig:pareto-throughput-coco}
\end{figure}

Now that we have established \name as a fair comparison tool, we can utilize it to measure segmentation efficiency across backbones.
All backbones in this section are pretrained on ImageNet-21k using the same DeiT~III setup.
As visualized in \Cref{fig:backbones-pretrain-vs-iou}, at the $224$\,px pre-training resolution, most mixers transfer to dense prediction without difficulty
while a distinct set does not:
ResNet-50~\cite{He2016}, CoaT~\cite{Xu2021}, and NextViT~\cite{Li2022b} (in throughput), which all utilize convolutions, as well as Switch~\cite{Fedus2022} are below the mean-transfer curve, contrasting BoB~\cite{Goldblum2023}, who found supervised CNNs to be the best backbones.
Note that ResT is the exception to this rule and that ResNet-50 and CoaT generalize better to Cityscapes, but not to panoptic segmentation (see \Cref{fig:pretrain-acc} in \Cref{apdx:bench-plots-backbones}).
NextViT is Pareto-optimal for memory-efficiency.
ViT, FocalNet~\cite{Yang2022a}, and WaveViT generalize well from classification to both segmentation tasks.

Transferring at $224$\,px does not imply transferring at the resolutions segmentation actually demands.
Across backbones, ImageNet top-1 accuracy is an unreliable predictor of segmentation performance at higher resolutions with Pearson $r \approx 0.72$ for ADE20K panoptic at $512$\,px and $r \approx 0.52$ for Cityscapes semantic at $1024$\,px with significant outliers to the top and bottom.
In particular, Kernel, Low-Rank, and Fixed attention all have problems generalizing to higher resolutions:
In \Cref{fig:pareto-memory-ade}\footnote{Plots with throughput and memory against ADE20K and Cityscapes semantic segmentation (\Cref{fig:eff-ade-sem,fig:eff-city-sem}), and COCO and ADE20K panoptic segmentation (\Cref{fig:eff-coco-pano,fig:eff-ade-pano}) quality are in \Cref{apdx:bench-plots-backbones}.}, Linformer drops from $38.9$ to $4.9$\,mIoU, Hydra from $34.9$ to $7.5$, and Synthesizer~\cite{Tay2021} from $37.0$ to $29.0$, inverting the $+4$--$6$\,mIoU gain that ViT and other backbones obtain.
Nystrom~\cite{Xiong2021} and PolySA increase performance only up to intermediate resolution.

In the throughput--PQ trade-off (\Cref{fig:pareto-throughput-coco} and \Cref{apdx:bench-plots-backbones}), even \emph{without FlashAttention}~\cite{Dao2022,Dao2024}, plain ViT still holds the Pareto frontier; the non-ViT models that come closest are sparse attention models (Swin-S~\cite{Liu2021} and WaveViT~\cite{Yao2022}), and FocalNet.
Enabling FlashAttention roughly doubles ViT's throughput, displacing other models from the throughput--frontier.
This exposes an implementation confound rather than an architectural difference:
Efficient mixers are measured with unoptimized kernels, and a comparably engineered implementation (e.g.\ a fused WaveViT) is the natural route to an efficient backbone that can better rival ViT with FlashAttention, as indicated by results relative to ViT \emph{without} FlashAttention.
For faster models, ResT~\cite{Zhang2021}, and HiViT~\cite{Zhang2023c} strike a Pareto-optimal balance.

\begin{takeaway}[Which backbone should I use?]
	A ViT backbone with FlashAttention is a solid choice (generalization and speed), with sparse attention, or FocalNet as efficient alternatives, or NextViT for memory.
\end{takeaway}

Note however, that differences are generally pretty small for the best ViT, hybrid, sparse, and non-attention models, especially for \emph{semantic} segmentation.

\subsection{Which pretraining to use?}
\label{sec:pretraining}
\begin{figure}
	\centering
	\includegraphics[width=0.95\columnwidth]{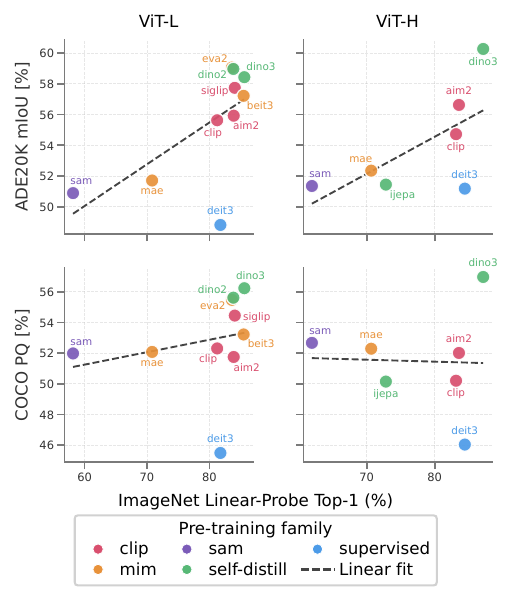}
	\caption{\textbf{Pretraining objective, not ImageNet probe, drives segmentation quality.} mIoU/PQ vs.\ ImageNet linear-probe (top: ADE20K semantic @$512$\,px, bottom: COCO panoptic @$640$\,px). Dense objectives perform well, supervised DeiT~III is the worst.}\label{fig:pretraining}
\end{figure}

With the architecture fixed, the next question is which pretraining to initialize it with.
The natural instinct is to pick the checkpoint based on the ImageNet linear-probe accuracy, since most papers report it.
\Cref{fig:pretraining} plots mIoU/PQ against ImageNet accuracy across 11 pretraining schemes under our unified recipe.
A positive trend exists only on ADE20K but is weak and carried almost entirely by the extremes, as removing plain MAE~\cite{He2022}, SAM~\cite{Kirillov2023}, and supervised DeiT~III~\cite{Touvron2022} collapses the rank correlation among the remaining strong models to $\rho = 0.21$ for ViT-L on ADE20K.
On Cityscapes and COCO for ViT-H, the trend is even negative\footnote{Pretraining evaluation plots for ADE20K and Cityscapes semantic and COCO and ADE20K panoptic segmentation are in \Cref{apdx:more-plots-pretrain}.}.
The dominant signal is instead the pretraining \emph{objective}.
Dense objectives (masked image modeling, especially EVA-02~\cite{Fang2024}, and DINO's~\cite{Oquab2024, Simeoni2025} self-distillation) meet or exceed the trend regardless of where they fall on ImageNet linear-probe, with EVA-02 the strongest model overall for ViT-L on ADE20K semantic (59.1\,mIoU) and panoptic segmentation (49.7\,PQ) despite only mid-pack classification accuracy.
On Cityscapes semantic and COCO panoptic segmentation, DINOv3 is the best pretraining, closely followed by DINOv2 (for ViT-L).
As already found by \citet{Kerssies2024}, SAM's~\cite{Kirillov2023} segmentation pretraining does \emph{not} generalize to our setup on ADE20K and COCO, but performs very well on Cityscapes (only DINO is better).
Since supervised and CLIP-like pretraining decouples global, linearly-separable semantics from the per-patch spatial structure, it does not generalize well to dense tasks:
DeiT~III's linear-probe (81.8\% on mean patch-embedding) exceeds both CLIP~\cite{Radford2021} and plain MAE, yet it sits $\approx 8$\,mIoU/PQ below the trend prediction.
CLIP itself also sits below the trend.
We conclude that self-supervised approaches have improved a lot since BoB~\cite{Goldblum2023}, when supervised pretraining was found best.
Our practical recommendation is:

\begin{takeaway}[How to pretrain for segmentation?]
	Initialize from a dense-objective foundation model (DINOv3 or EVA-02 work best) and disregard ImageNet linear-probe accuracy as a proxy.
\end{takeaway}

Note, that \Cref{sec:backbone} holds pretraining fixed to a shared supervised pretraining to ensure a controlled test of transfer from identical conditions, since dense-objective pretrained states are not (yet) available for most architectures.

\subsection{Compute-Optimal Scaling}
\label{sec:scaling}

\begin{figure}
	\includegraphics[width=0.95\columnwidth]{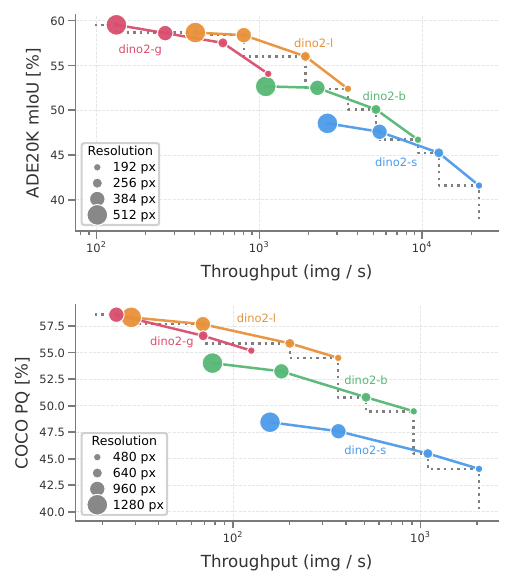}
	\caption{\textbf{Compute-optimal scaling using DINOv2.} Throughput vs.\ segmentation quality on ADE20K semantic mIoU (top) and COCO panoptic PQ (bottom) as model size and resolution vary. The compute-optimal frontier (\Cref{eq:surface}) directs roughly $80\%$ of added compute to model size.}\label{fig:model-res-scaling}
\end{figure}

A fixed compute budget can be spent on a larger backbone or on higher input resolution.
Both raise segmentation quality, but which is more efficient?
We characterize the optimal trade-off by comparing DINOv2~\cite{Oquab2024} across backbone sizes and input resolutions under our unified recipe (see \Cref{fig:model-res-scaling}), we find that the Pareto-optimal models are identical whether cost is measured in FLOPs, inference memory, or latency, indicating a general allocation rule.
Accuracy surfaces are well described by a saturating power law, mirroring the functional form used to model joint parameter--data scaling in language models~\cite{Hoffmann2022} and compute scaling in vision~\cite{Zhai2022}:
\begin{equation}
	\mathrm{mIoU}(N, R) \;=\; U - A\,N^{-\alpha} - B\,R^{-\beta},
	\label{eq:surface}
\end{equation}
with parameter count $N$ and resolution $R$, fitting to within $0.3$\,mIoU on Cityscapes and $0.7$ on ADE20K.
Since inference cost follows $C \propto N^{p} R^{q}$ with $p=0.97$, $q=2.22$ ($R^2=0.999$), minimizing \Cref{eq:surface} under a budget $C$ yields closed-form compute-optimal trajectories $N^\star \propto C^{e_N}$ and $R^\star \propto C^{e_R}$ (derivation in \Cref{apdx:optimal-scaling}).

We obtain $e_N = 0.8$ on ADE20K (panoptic), Cityscapes, and COCO, against $e_R = 0.1$, but $e_N = 0.96$ and $e_R = 0.03$ on ADE20K (semantic), hinting at an even stronger model size scaling for that task.
Note, that we fit our data to DINOv2-S/-B/-L only, since these are all distilled from DINOv2-g and thus should have similar pretraining quality.
This is also supported by \Cref{fig:model-res-scaling}, where the DINOv2\nobreakdash-g curve falls clearly \emph{below} the DINOv2-L one at similar throughput.
Scaling the model alone at fixed resolution turns a $10\times$ budget into two ViT size steps (S$\to$L); the optimal path (for $e_N=0.8$, $e_R=0.1$) instead diverts $20$--$25$\% of the additional compute to resolution while buying a $\approx 5$--$7\times$ larger backbone.
All in all, the optimal scaling, given similar pretraining quality across model sizes, is:

\begin{takeaway}[How to scale up efficiently?]
	Roughly $80\%$ of additional compute should be spent on model size: $N^\star \propto C^{0.8}$, with optimal token count growing only as $T^\star \propto C^{0.2}$.
\end{takeaway}

\section{Conclusion}

We ask which backbone is most efficient for segmentation.
Answering it required an instrument the literature lacks: a single lightweight head that attaches unchanged to any token mixer, so that a measured difference reflects the backbone and not the apparatus around it.
\name is that instrument, and it shows that EoMT's~\cite{Kerssies2025} performance and efficiency can be matched by a minimal external read.

Holding \name fixed across 20 backbones, 11 pretraining schemes, and a range of resolutions separates, for the first time under one modern recipe, what governs semantic and panoptic segmentation quality.
Along the \emph{architecture} axis the differences largely collapse:
Good backbones from multiple families transfer within a narrow band, and the novel token-mixer designs we test deliver neither the accuracy nor the efficiency they promise, even with high-resolution inputs.
The plain ViT holds the throughput--front, with efficient variants competitive only in memory.
A few sequence-length-bound mixers collapse outright as resolution grows.
Along the \emph{pretraining} axis the task is important:
As \citet{Kerssies2024} also found, dense objectives transfer robustly, while supervised and multimodal contrastive pretraining fall systematically below the trend compared to classification accuracy.
Progress in self-supervised pretraining since BoB~\cite{Goldblum2023} suggests opportunities for further gains here.
Along the \emph{scale} axis the compute-optimal frontier is dominated by model size, which should absorb roughly $80\%$ of any added budget.
Architecture, the lever the field has tuned hardest, moves dense performance only little; the pretraining objective and plain model scaling move it most.

The thread through all three is that the proxies ranking models at classification---top-1 accuracy, linear-probe separability---can mislead for segmentation, because they reward globally separable semantics over the per-patch spatial structure dense prediction demands.
The practical synthesis is therefore simple:
Keep a simple architecture, initialize from a dense-objective foundation model, and spend added compute on model size before resolution.
All in all, effort is best directed not at new token mixers, but at pretraining objectives and scalable architectures measured fairly on the dense task itself, which \name now makes possible.

\subsection*{Acknowledgements}
This work was funded by the Carl-Zeiss Foundation under the Sustainable Embedded AI project (P2021-02-009) and by the BMFTR project Albatross (funding code 16IW24002).
All compute was done on the Pegasus cluster at DFKI Kaiserslautern.

{
	\small
	\bibliographystyle{ieeenat_fullname}
	\bibliography{main}
}

\newpage
\appendix
\onecolumn
\section{Compute-optimal allocation: derivation}
\label{apdx:optimal-scaling}

We use the size--resolution sweep of \Cref{sec:scaling} to fit our scaling laws: ViT-\{S, B, L, g\} with DINOv2~\cite{Oquab2024} pretraining, trained with \name{} at $\{192, 256, 384, 512\}$\,px on ADE20K~\cite{Zhou2019} semantic, $\{480, 640, 960, 1280\}$\,px on ADE20K panoptic and on COCO~\cite{Lin2014_Microsoft_COCO_Comm}, and $\{384, 512, 768, 1024\}$\,px on Cityscapes~\cite{Cordts2016}.
All runs share the head configuration, patch size, optimizer, and schedule and each configuration is trained at its target resolution.
Both mIoU and PQ are computed at full ground-truth resolution, so scores are comparable across input resolutions.

We fit $\log C = \log k + p \log N + q \log R$ on measured inference FLOPs, separately per task: $p = 0.97$, $q = 2.18$ on ADE20K semantic; $p = 0.90$, $q = 2.76$ on ADE20K panoptic; $p = 0.91$, $q = 2.60$ on Cityscapes; and $p = 0.90$, $q = 2.76$ on COCO, each with $R^2 \geq 0.99$.
We fit the accuracy surface (\Cref{eq:surface}) per task by nonlinear least squares (multi-start, $U$ bounded above by 100).
The two error terms are additive and separable, which the data supports: residuals show no $N \times R$ interaction structure beyond the ViT-g deviation discussed below.
\Cref{tab:scaling_fits} reports all fits.
We report the S/B/L fit as primary on all four tasks.
ViT-g falls below the surface the smaller sizes define everywhere we fit it---by up to $2.1$\,mIoU on Cityscapes, $3.5$\,mIoU on ADE20K semantic, $3.6$\,PQ on COCO, and $5.4$\,PQ on ADE20K panoptic---which we read as a property of the DINOv2 family rather than of segmentation.
ViT-g/14 is trained from scratch while S/B/L are distilled from it, so the S$\to$B$\to$L trend is coming from distillation of the same teacher.
The all-sizes fits are reported alongside as a sensitivity.

\begin{table}[h]
	\centering
	\caption{\textbf{Power-law fits and resulting compute-optimal exponents.} Brackets give $68\%$ bootstrap intervals over runs. $^\dagger$Ceiling at its bound ($U{=}100$): the $U$--$\alpha$ degeneracy at a $13\times$ parameter span makes $\alpha$ unreliable in isolation, while the exponents $e_N, e_R$ remain the stable quantities. Primary fits in bold.}
	\label{tab:scaling_fits}
	\begin{tabular}{llcccccc}
		\toprule
		Task                     & Sizes          & $n$ & $\alpha$       & $\beta$ & RMSE & $e_N$                                   & $e_R$                                   \\
		\midrule
		\textbf{ADE20K sem.}     & \textbf{S/B/L} & 12  & 0.09$^\dagger$ & 2.53    & 0.61 & \textbf{0.96} {\scriptsize[0.93, 0.97]} & \textbf{0.03} {\scriptsize[0.03, 0.05]} \\
		ADE20K sem.              & S/B/L/g        & 16  & 0.27           & 2.63    & 0.89 & 0.84 {\scriptsize[0.73, 0.89]}          & 0.09 {\scriptsize[0.06, 0.14]}          \\
		\midrule
		\textbf{ADE20K pan.}     & \textbf{S/B/L} & 11  & 0.11$^\dagger$ & 1.01    & 0.27 & \textbf{0.83} {\scriptsize[0.70, 0.92]} & \textbf{0.09} {\scriptsize[0.06, 0.13]} \\
		ADE20K pan.              & S/B/L/g        & 13  & 0.39           & 1.57    & 0.91 & 0.63 {\scriptsize[0.10, 0.80]}          & 0.16 {\scriptsize[0.10, 0.33]}          \\
		\midrule
		\textbf{Cityscapes sem.} & \textbf{S/B/L} & 10  & 0.14$^\dagger$ & 1.03    & 0.29 & \textbf{0.80} {\scriptsize[0.66, 0.87]} & \textbf{0.11} {\scriptsize[0.08, 0.16]} \\
		Cityscapes sem.          & S/B/L/g        & 13  & 0.35           & 0.84    & 0.43 & 0.50 {\scriptsize[0.36, 0.67]}          & 0.21 {\scriptsize[0.15, 0.26]}          \\
		\midrule
		\textbf{COCO pan.}       & \textbf{S/B/L} & 11  & 0.10$^\dagger$ & 0.78    & 0.23 & \textbf{0.81} {\scriptsize[0.69, 0.90]} & \textbf{0.10} {\scriptsize[0.07, 0.14]} \\
		COCO pan.                & S/B/L/g        & 13  & 0.38           & 0.89    & 0.63 & 0.49 {\scriptsize[0.11, 0.73]}          & 0.20 {\scriptsize[0.12, 0.33]}          \\
		\bottomrule
	\end{tabular}
\end{table}
Minimizing $E(N, R) = A N^{-\alpha} + B R^{-\beta}$ subject to $k N^{p} R^{q} = C$, stationarity of the Lagrangian in log-coordinates gives
\begin{equation}
	\frac{\alpha A N^{-\alpha}}{p} \;=\; \frac{\beta B R^{-\beta}}{q},
	\label{eq:stationarity}
\end{equation}
i.e.\ the marginal error reduction per unit of log-compute is equalized across the two levers.
\Cref{eq:stationarity} implies $N^{-\alpha} \propto R^{-\beta}$ along the optimal path; differentiating this relation together with the constraint $p\,\mathrm{d}\!\log N + q\,\mathrm{d}\!\log R = \mathrm{d}\!\log C$ yields
\begin{equation}
	e_N = \frac{\beta}{p\beta + q\alpha},
	\qquad
	e_R = \frac{\alpha}{p\beta + q\alpha},
	\qquad
	E^\star \propto C^{-\frac{\alpha\beta}{p\beta + q\alpha}},
	\label{eq:exponents}
\end{equation}
which satisfy $p\,e_N + q\,e_R = 1$ by construction.
Numerically, the optimal error decays as $E^\star \propto C^{-0.08}$ on ADE20K semantic, $C^{-0.09}$ on ADE20K panoptic, $C^{-0.11}$ on Cityscapes, and $C^{-0.08}$ on COCO.
Confidence intervals in \Cref{tab:scaling_fits} are obtained by refitting on 400 bootstrap resamples of the runs and propagating through \Cref{eq:exponents}.
Note, that this allocation holds only where the pretraining method can effectively pretrain the larger model's weights. ViT-g is the counterexample in our own data, so the exponents say how to spend compute given pretraining that scales with model size, not that a larger backbone pays off however it was pretrained.

\smallskip
\noindent
Two limitations bound these estimates:
\begin{itemize}
	\item Every sweep reaches or passes the dataset's native image resolution at its largest setting, so $R^\star$ extrapolations beyond it are meaningless.
	\item Excluding ViT-g restricts every primary fit to a $13\times$ parameter span across three sizes, which widens the $e_N$ interval and creates the $U$--$\alpha$ degeneracy noted in \Cref{tab:scaling_fits}.
\end{itemize}

\section{Hyperparameters \& Setup}
\label{apdx:setup}

We describe our setup for training and evaluation using \name. For more details, also see our codebase\footnote{\code{https://github.com/tobna/LUMA}}.

\subsection{Training Setup.}
Every backbone in the benchmark is trained with the identical recipe below; the only quantities that vary per backbone are the pretrained checkpoint, and the patch size.
No per-backbone hyperparameter tuning is performed, so that any difference in mIoU or PQ is attributable to the architecture rather than to optimization.

We optimize with AdamW at a base learning rate of $1{\times}10^{-4}$ and weight decay $0.05$, in \texttt{bf16-mixed} precision.
LLRD~\cite{Bao2022} with factor $0.8$ is applied across the backbone, the pretrained non-block parameters (patch and positional embeddings) take the maximum discount $\text{lr}\times 0.8^{\,N-1}$, and the \name{} queries, heads, and upscale modules train at the full base rate.
The backbone is fine-tuned, not frozen: a two-stage warmup holds it at zero learning rate for the first $500$ steps while the new \name{} modules warm up linearly over the same window, after which the backbone ramps linearly over $1000$ steps; both groups then follow cosine decay to zero.
We inject interleaved cross-attention into the last $L_2=2$ backbone blocks, with $100$ learnable queries for semantic and $200$ for panoptic segmentation, following EoMT~\cite{Kerssies2025}.
On the panoptic tasks the cross-attention is additionally masked by the mask prediction of the preceding block, and the mask is annealed to uniform attention.
We leave the masking off on the semantic tasks, as it is not needed.

We adopt the Hungarian-matched Mask2Former loss~\cite{Cheng2022}: cross-entropy on the class logits (weight $2.0$, with $\varnothing$-class weight $0.1$), and binary cross-entropy (weight $5.0$) plus Dice loss (weight $5.0$) on the mask logits.
The mask loss is computed on $12544$ points sampled with oversample ratio $3.0$ and importance-sample ratio $0.75$.
This is unchanged across all four tasks.

Augmentation is random horizontal flip, scale jitter, random crop to the training resolution, and color jitter (brightness $\pm 32/255$, contrast and saturation $\pm 0.5$, hue $\pm 18^\circ$).
The semantic tasks jitter the scale in $(0.5, 2.0)$; the panoptic tasks use the wider jitter range $(0.1, 2.0)$ of Mask2Former, and COCO is trained without color jitter.

\begin{table}[t]
	\centering
	\begin{tabular}{lcccc}
		\toprule
		\multirow{2.5}{*}{Setting}       & \multicolumn{2}{c}{Semantic} & \multicolumn{2}{c}{Panoptic}                                   \\
		\cmidrule(lr){2-3}\cmidrule(lr){4-5}
		                                 & ADE20K                       & Cityscapes                   & ADE20K         & COCO           \\
		\midrule
		Train resolution                 & $512\times512$               & $1024\times1024$             & $640\times640$ & $640\times640$ \\
		Classes                          & $150$                        & $19$                         & $100+50$       & $80+53$        \\
		Training images                  & $20\,210$                    & $2\,975$                     & $20\,210$      & $118\,287$     \\
		Queries                          & $100$                        & $100$                        & $200$          & $200$          \\
		Masked cross-attention           & \xmark                       & \xmark                       & \cmark         & \cmark         \\
		Scale jitter                     & $(0.5, 2.0)$                 & $(0.5, 2.0)$                 & $(0.1, 2.0)$   & $(0.1, 2.0)$   \\
		Color jitter                     & \cmark                       & \cmark                       & \cmark         & \xmark         \\
		GPUs $\times$ batch size per GPU & $2\times 8$                  & $8\times 2$                  & $4\times 4$    & $4\times 4$    \\
		Effective batch size             & $16$                         & $16$                         & $16$           & $16$           \\
		Epochs                           & $31$                         & $107$                        & $31$           & $12$           \\
		\bottomrule
	\end{tabular}
	\caption{\textbf{Task-specific training settings.} All other hyperparameters are shared.
	For panoptic, classes are given as \textit{things}${}+{}$\textit{stuff}.
	The effective batch size is held at $16$ everywhere, so the split into GPUs and per-GPU batch size shifts to $8\times 2$ for the largest models and resolutions without changing the schedule.}
	\label{tab:appendix-dataset-settings}
\end{table}

The four tasks differ only in the quantities listed in \Cref{tab:appendix-dataset-settings}.
At validation the semantic tasks use sliding-window inference, so the image is resized such that its short side matches the training resolution, tiled with overlap, and the window logits are averaged.
The panoptic tasks instead resize the long side to the training resolution, pad to a square, and run a single forward pass; the mask and class logits are merged into a panoptic map with mask and overlap thresholds of $0.8$.

\subsection{Efficiency Evaluation.}
\label{apdx:eval-setup}
All efficiency numbers are measured on a single NVIDIA H100 in \texttt{bf16} with \texttt{torch.compile}.
FLOPs are computed with \code{fvcore}~\footnote{\url{https://github.com/facebookresearch/fvcore}}.
In \Cref{sec:luma-segmentation} in \Cref{tab:luma-semantic-seg} where \name{} is compared against external baselines, FLOPs and FPS account for the full sliding-window inference and FPS is reported at single-image (batch size $1$) latency, matching the comparison protocol of EoMT~\cite{Kerssies2025}.
The remaining sections report inference throughput at the optimal batch size and do not fold in the windowing factor, since that factor is constant across all \name{} variants and cancels in any cross-backbone comparison.

\begin{table}
	\centering
	\caption{\textbf{Run-to-run variance.} Mean score and standard deviation over repeated trainings of the same setting.
		The mean standard deviation is $0.49$\,mIoU on ADE20K, $0.11$\,mIoU on Cityscapes, and $0.29$\,PQ on COCO.
	}\label{tab:variance}
	\begin{tabular}{llcccccc}
  \toprule
  \multirow{2.5}{*}{Backbone} & \multirow{2.5}{*}{Pretraining} & \multicolumn{2}{c}{ADE20K (semantic)} & \multicolumn{2}{c}{Cityscapes (semantic)} & \multicolumn{2}{c}{COCO (panoptic)} \\
  \cmidrule(lr){3-4} \cmidrule(lr){5-6} \cmidrule(lr){7-8}
   & & mIoU\,(\%) & Std\,(\%) & mIoU\,(\%) & Std\,(\%) & PQ\,(\%) & Std\,(\%) \\
  \midrule
  ViT-L & DINOv2 & 58.41 & 0.18 & 84.65 & 0.01 & 55.73 & 0.13 \\
  ViT-L & DeiT III & 49.59 & 0.62 & — & — & 45.58 & 0.14 \\
  ViT-B & DeiT III & 49.34 & 0.54 & 78.64 & 0.04 & 47.37 & 0.13 \\
  FocalNet-S-Srf & DeiT III & 48.48 & 0.57 & 76.58 & 0.15 & 40.81 & 0.15 \\
  WaveViT-S & DeiT III & 46.57 & 0.54 & 76.46 & 0.24 & 38.92 & 0.89 \\
  \bottomrule
\end{tabular}

\end{table}

\section{Run-to-Run Variance}
\label{apdx:variance}

Since it is infeasible to run every one of our benchmark experiments multiple times, we verify run-to-run variance on a representative subset of configurations, re-trained with only the random seed varying.
The subset covers five backbone and pretraining pairs on ADE20K semantic, COCO panoptic, and Cityscapes semantic at three seeds each.
\Cref{tab:variance} reports the resulting spread.
The mean standard deviation is $0.49$\,mIoU on ADE20K and $0.29$\,PQ on COCO, and a good deal lower at $0.11$\,mIoU on Cityscapes.
On COCO, WaveViT-S has a standard deviation of $0.89$\,PQ, while the other models stay between $0.13$ and $0.15$.
Treating the floor as homoscedastic, two single runs of the same setting differ by more than $1.4$\,mIoU on ADE20K, $0.8$\,PQ on COCO, or $0.3$\,mIoU on Cityscapes in only $\sim$\,5\% of cases.
All comparisons on which our conclusions rest span several points of mIoU or PQ and therefore exceed this run-to-run noise by a wide enough margin, confirming that single-seed reporting is sufficient for the analysis, especially since trends replicate across tasks.

\section{Ablating \name}
\label{apdx:luma-ablation}

\begin{figure}
	\begin{subfigure}{.49\textwidth}
		\centering
		\includegraphics[width=0.99\columnwidth]{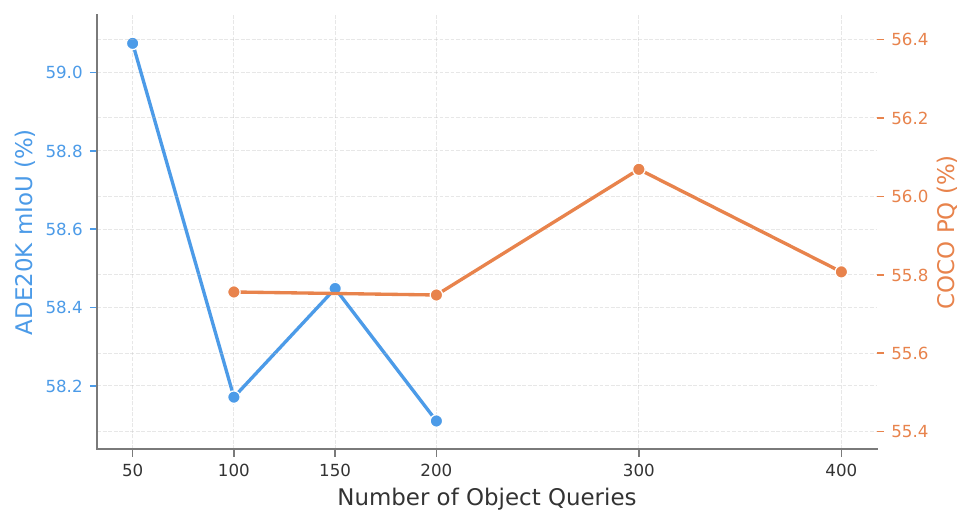}
	\end{subfigure}
	\hfill
	\begin{subfigure}{.49\textwidth}
		\centering
		\includegraphics[width=0.99\columnwidth]{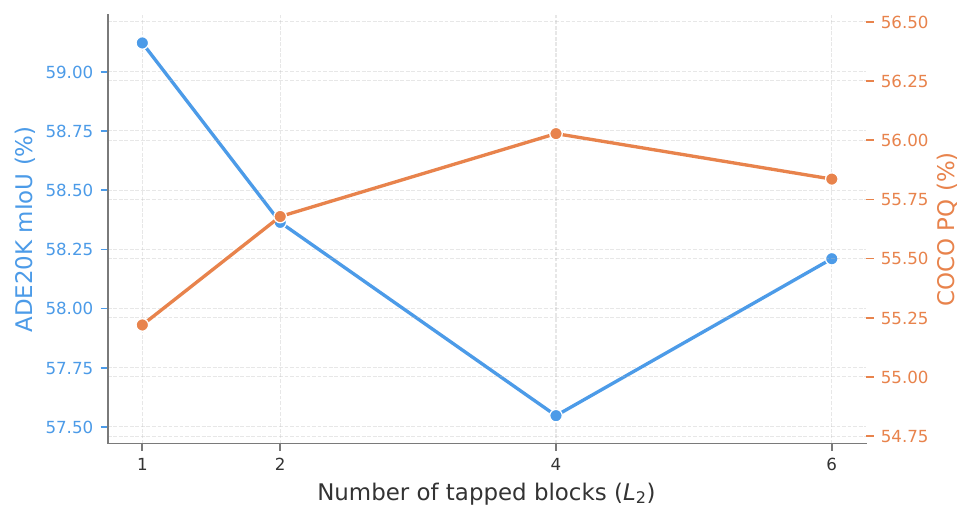}
	\end{subfigure}
	\caption{\textbf{Number of queries and tapped blocks ($L_2$).} DINOv2-L on ADE20K semantic and COCO panoptic. \textbf{Left:} Queries. For our main experiments, we use EoMT's settings of $100$ queries for semantic and $200$ queries for panoptic segmentation. \textbf{Right:} Tapped blocks. More tapped blocks \emph{cost} mIoU on semantic segmentation but \emph{gain} PQ on panoptic segmentation, which is why our unified recipe settles on $L_2 = 2$.}\label{fig:ablation}
\end{figure}

\begin{table}
	\centering
	\caption{\textbf{Decoder settings.} DINOv2-L with $L_2 = 2$, one setting changed at a time; the shaded row is our default, and the best value per block and column is bold. Masking is only used on the panoptic tasks.}\label{tab:ablation}
	\begin{tabular}{lcc}
  \toprule
  Setting & ADE20K mIoU\,(\%) & COCO PQ\,(\%) \\
  \midrule
  \multicolumn{3}{l}{\textit{LayerScale init $\gamma_0$}} \\[1pt]
  $10^{-5}$ & 58.3 & 55.6 \\
  \rowhighlight $1$ & \textbf{58.4} & \textbf{55.7} \\
  \midrule
  \multicolumn{3}{l}{\textit{Query MLP}} \\[1pt]
  None & 58.7 & 55.5 \\
  \rowhighlight Shared with backbone & 58.4 & \textbf{55.7} \\
  Separate & \textbf{58.8} & 55.6 \\
  \midrule
  \multicolumn{3}{l}{\textit{Query self-attention}} \\[1pt]
  Off & 58.2 & 54.9 \\
  \rowhighlight On & \textbf{58.4} & \textbf{55.7} \\
  \midrule
  \multicolumn{3}{l}{\textit{Masked cross-attention}} \\[1pt]
  Off & --- & 55.4 \\
  \rowhighlight On & --- & \textbf{55.7} \\
  \bottomrule
\end{tabular}

\end{table}

The query count does not have much influence (\Cref{fig:ablation}, left).
ADE20K stays within $58.1$--$58.5$\,mIoU from $100$ to $200$ queries, COCO within $55.8$--$56.1$\,PQ from $100$ to $400$.
While on ADE20K $50$ queries are best, we keep EoMT's defaults of $100$ queries for semantic and $200$ for panoptic segmentation.

The number of blocks ($L_2$; \Cref{fig:ablation}, right) has a bigger influence and clearer trends.
On ADE20K semantic segmentation, \name is best with a single cross-attention read at $59.1$\,mIoU and loses $1.5$\,mIoU by $L_2 = 4$, while COCO panoptic segmentation gains $0.8$\,PQ over the same range.
Since we want one recipe for all four tasks, we use $L_2 = 2$, which costs $0.8$\,mIoU and $0.3$\,PQ against the per-task optimum.

\Cref{tab:ablation} ablates \name's remaining settings.
The LayerScale init and both query MLP variants stay within $0.4$\,mIoU and $0.2$\,PQ of the default.
We pick to share the MLP with the backbone by default to save on parameters.
For models where that is not possible, like ResNet, we fall back to adding a separate MLP.
Adding query self-attention is the biggest gain for panoptic segmentation with $+0.8$\,PQ.
It acts mostly on \emph{things}, which gain $1.4$\,PQ while \emph{stuff} loses $0.3$\,PQ, with the gain sitting in recognition quality ($+1.9$\,RQ on things), so the queries need to attend to each other to keep instances apart.
Using annealed cross-attention masking aids training of COCO panoptic segmentation ($+0.3$\,PQ).

\section{More Benchmark Plots}
\label{apdx:bench-plots}

The Cityscapes runs in this section use a different \name{} configuration than the other three tasks: one tapped block ($L_2 = 1$) and no self-attention between the queries.
Cityscapes scores are therefore comparable among themselves, but a Cityscapes-to-ADE20K or Cityscapes-to-COCO gap also contains the decoder change.
The Cityscapes configs in \Cref{tab:luma-semantic-seg} of the main paper use the same \name{} config as the other tasks.

\subsection{Backbones}
\label{apdx:bench-plots-backbones}

\begin{figure}
	\begin{subfigure}{.49\textwidth}
		\centering
		\includegraphics[width=.90\textwidth]{fig/pretrain_acc_vs_score_ade_sem.pdf}
		\caption{ADE20K mIoU at $224$\,px.}
	\end{subfigure}
	\hfill
	\begin{subfigure}{.49\textwidth}
		\centering
		\includegraphics[width=.90\textwidth]{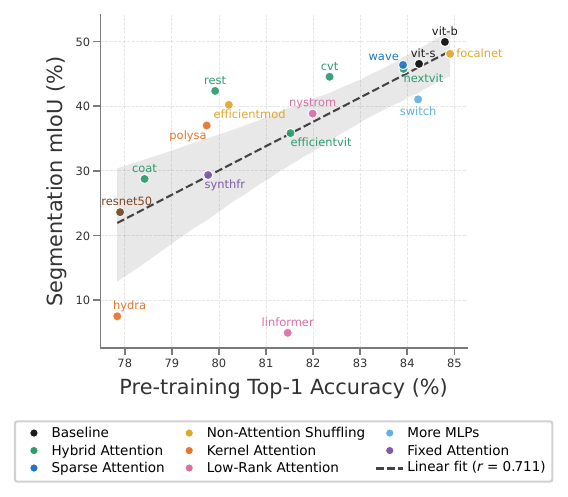}
		\caption{ADE20K mIoU at $512$\,px.}
	\end{subfigure}

	\begin{subfigure}{.49\textwidth}
		\centering
		\includegraphics[width=.90\textwidth]{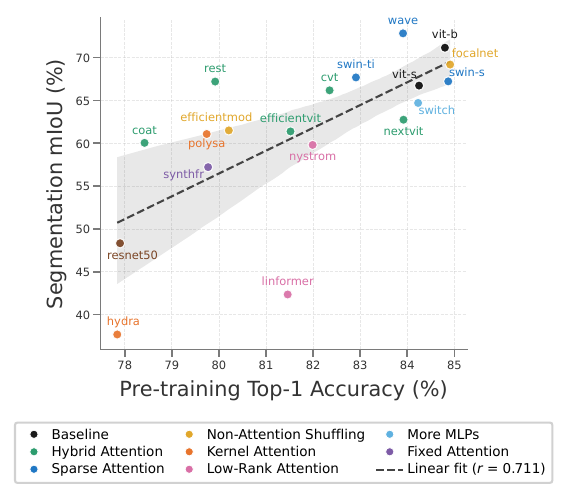}
		\caption{Cityscapes mIoU at $448$\,px.}
	\end{subfigure}
	\hfill
	\begin{subfigure}{.49\textwidth}
		\centering
		\includegraphics[width=.90\textwidth]{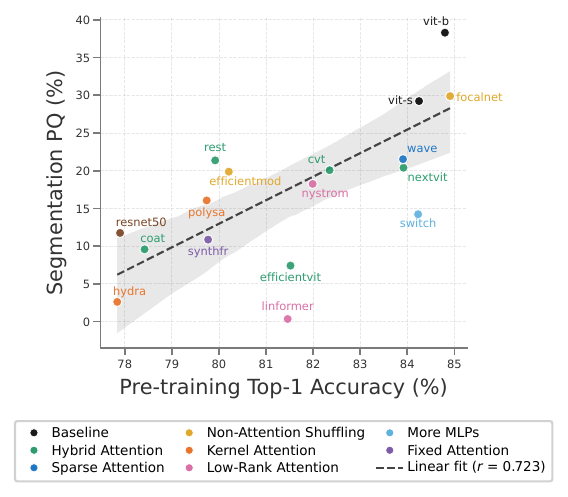}
		\caption{ADE20K PQ at $640$\,px.}
	\end{subfigure}

	\begin{subfigure}{.49\textwidth}
		\centering
		\includegraphics[width=.90\textwidth]{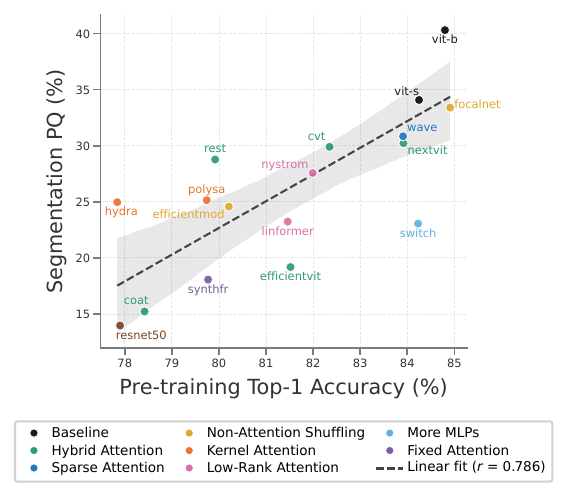}
		\caption{COCO PQ at $320$\,px.}
	\end{subfigure}
	\hfill
	\begin{subfigure}{.49\textwidth}
		\centering
		\includegraphics[width=.90\textwidth]{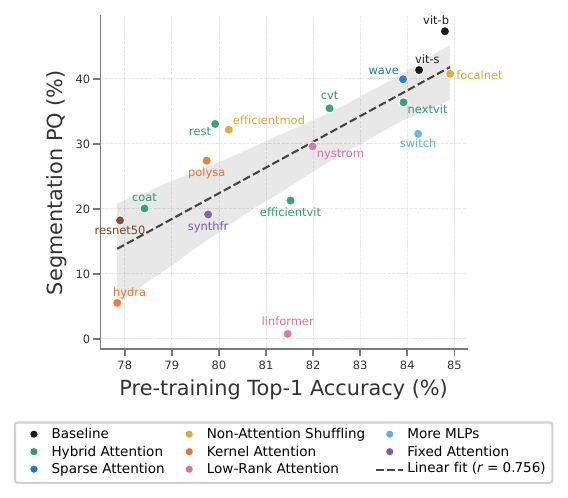}
		\caption{COCO PQ at $640$\,px.}
	\end{subfigure}
	\caption{\textbf{ImageNet top-1 against segmentation quality.}
		Cross-task correlation generally worsens and more extreme outliers appear with increased resolution. All models were pretrained at $224$\,px.
		CoaT and ResNet-50 both do better against the fit on Cityscapes than on ADE20K semantic, but recover performance at higher resolutions.}
	\label{fig:pretrain-acc}
\end{figure}

The fit ranges from $r = 0.71$ on ADE20K semantic at $512$\,px to $0.79$ on COCO (\Cref{fig:pretrain-acc}), similar to the $0.80$ of ADE20K semantic at $224$\,px (the pretraining resolution).
The backbones that move most from the fit when raising the resolution are the sequence-length-bound mixers.

\begin{figure}
	\begin{subfigure}{.49\textwidth}
		\centering
		\includegraphics[width=.90\textwidth]{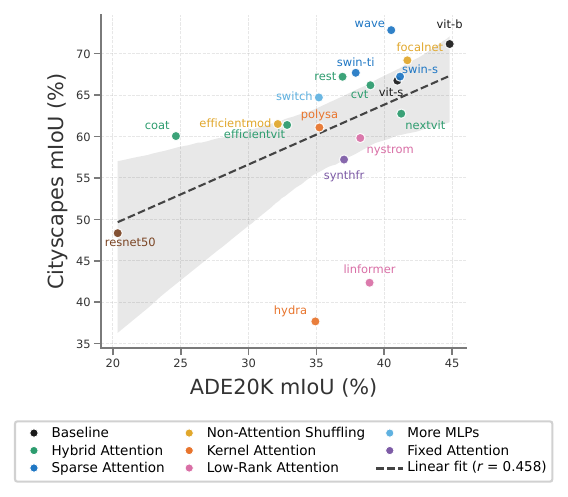}
		\caption{ADE20K vs.\ Cityscapes, both semantic.}
	\end{subfigure}
	\hfill
	\begin{subfigure}{.49\textwidth}
		\centering
		\includegraphics[width=.90\textwidth]{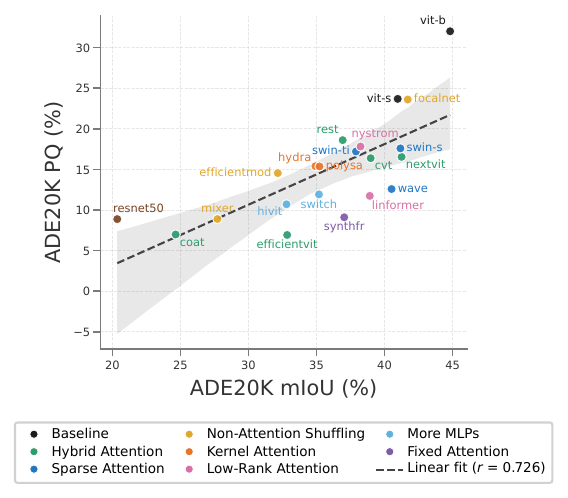}
		\caption{ADE20K semantic vs.\ ADE20K panoptic.}
	\end{subfigure}

	\begin{subfigure}{.49\textwidth}
		\centering
		\includegraphics[width=.90\textwidth]{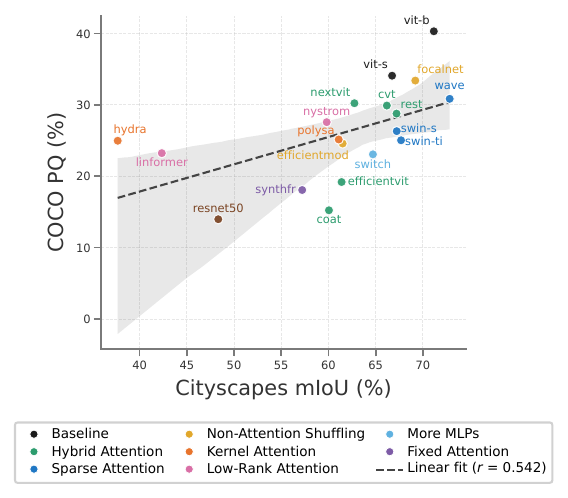}
		\caption{Cityscapes semantic vs.\ COCO panoptic.}
	\end{subfigure}
	\hfill
	\begin{subfigure}{.49\textwidth}
		\centering
		\includegraphics[width=.90\textwidth]{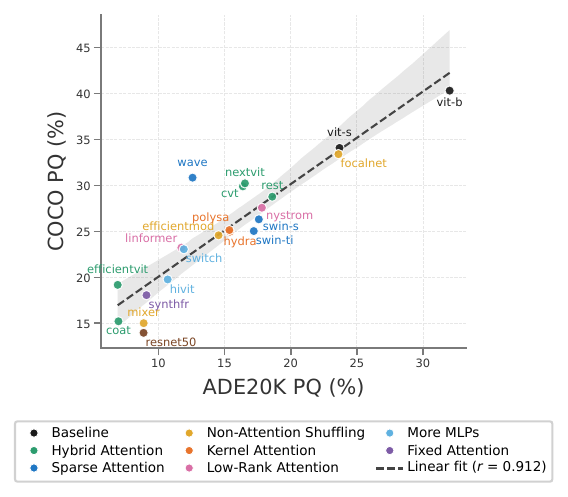}
		\caption{ADE20K vs.\ COCO, both panoptic.}
	\end{subfigure}
	\caption{\textbf{Benchmarks disagree with each other.}
		Each backbone at its lowest evaluated resolution on each task.
		ADE20K and Cityscapes correlate at $r = 0.46$, less than either correlates with ImageNet accuracy; note, that this includes a change of \name{} configuration on top of the dataset change (see the note above).
		The panoptic segmentation tasks are highly correlated at $r = 0.91$, while semantic-to-panoptic correlation is lower:
		$r = 0.73$ on the same ADE20K dataset and $r = 0.54$ across datasets. }
	\label{fig:cross-task}
\end{figure}

\begin{figure}
	\begin{subfigure}{\textwidth}
		\centering
		\includegraphics[width=.80\textwidth]{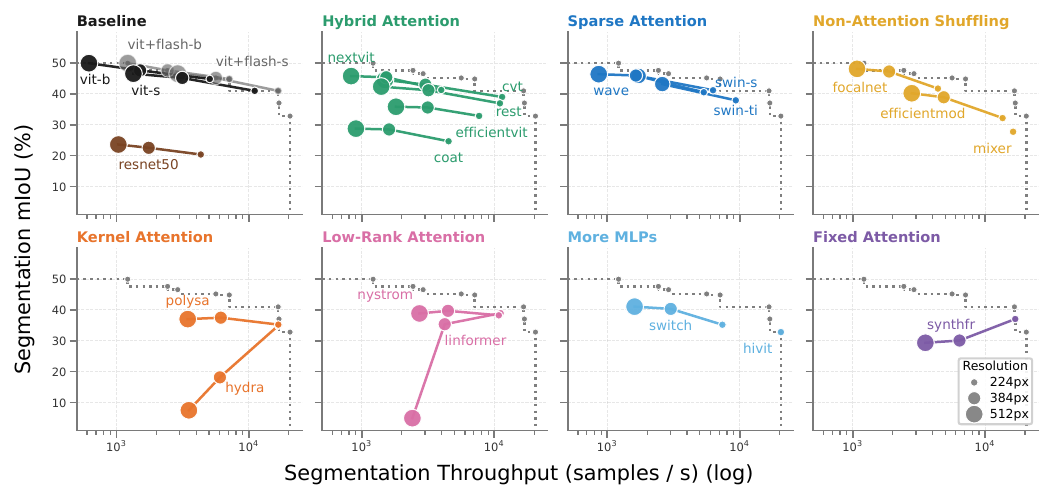}
		\caption{Throughput vs.\ ADE20K mIoU.}
	\end{subfigure}
	\begin{subfigure}{\textwidth}
		\centering
		\includegraphics[width=.80\textwidth]{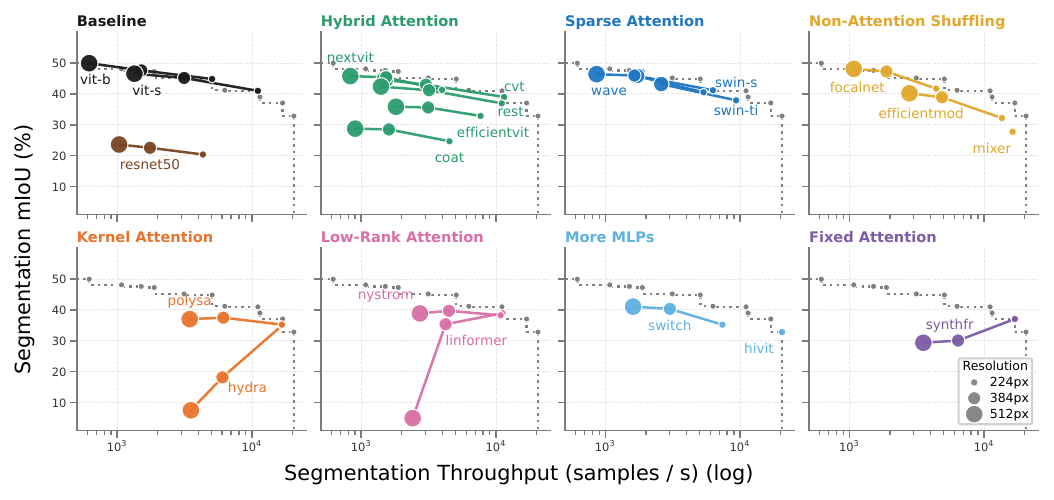}
		\caption{Throughput vs.\ ADE20K mIoU \textbf{without FlashAttention}.}
	\end{subfigure}
	\begin{subfigure}{\textwidth}
		\centering
		\includegraphics[width=.80\textwidth]{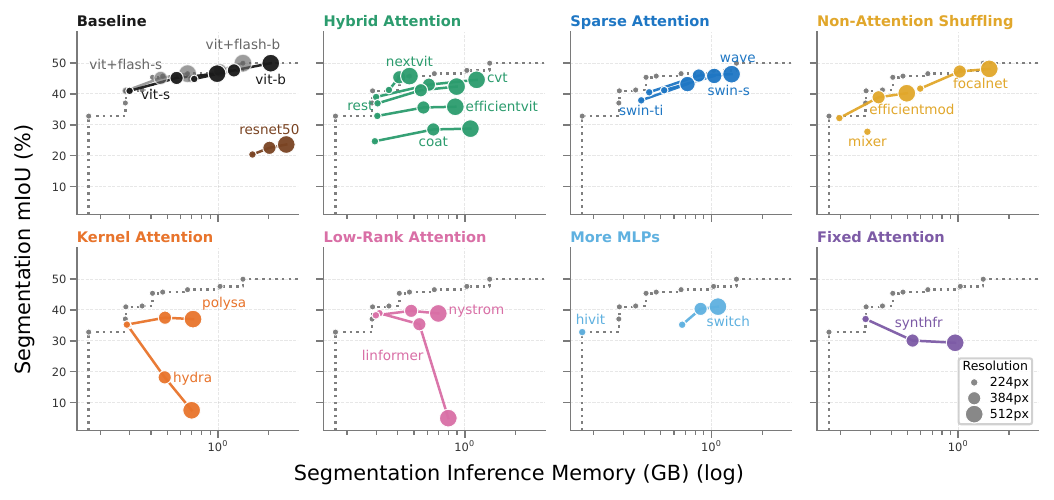}
		\caption{Inference memory vs.\ ADE20K mIoU.}
	\end{subfigure}
	\caption{\textbf{Efficiency on ADE20K semantic segmentation.}
		The dotted staircase is the Pareto front over the whole benchmark.
		ViT with FlashAttention holds both frontiers; among the efficient backbones only sparse attention (Swin-S, WaveViT) and non-attention shuffling (FocalNet) come close.
		Kernel, low-rank, and fixed mixers leave the front as resolution grows.}
	\label{fig:eff-ade-sem}
\end{figure}

\begin{figure}
	\begin{subfigure}{\textwidth}
		\centering
		\includegraphics[width=.8\textwidth]{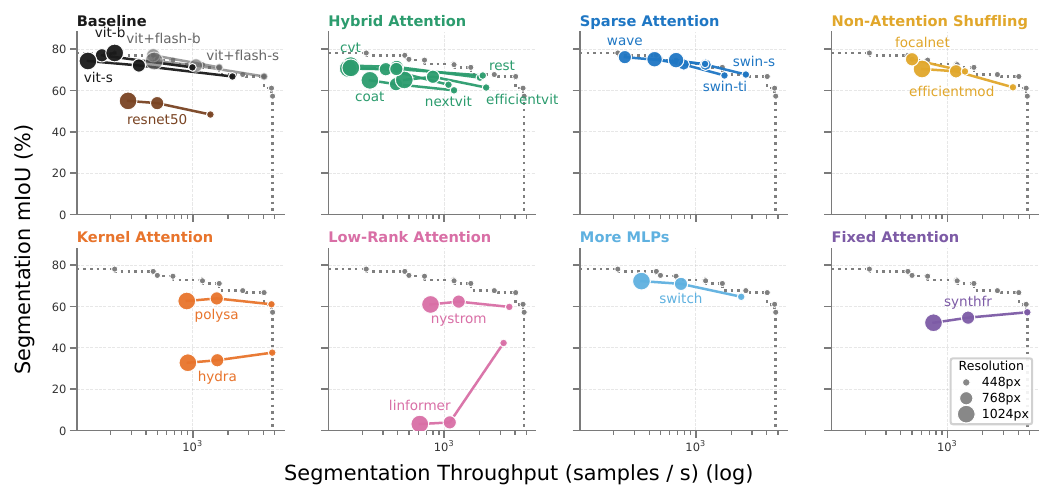}
		\caption{Throughput vs.\ Cityscapes mIoU.}
	\end{subfigure}
	\begin{subfigure}{\textwidth}
		\centering
		\includegraphics[width=.8\textwidth]{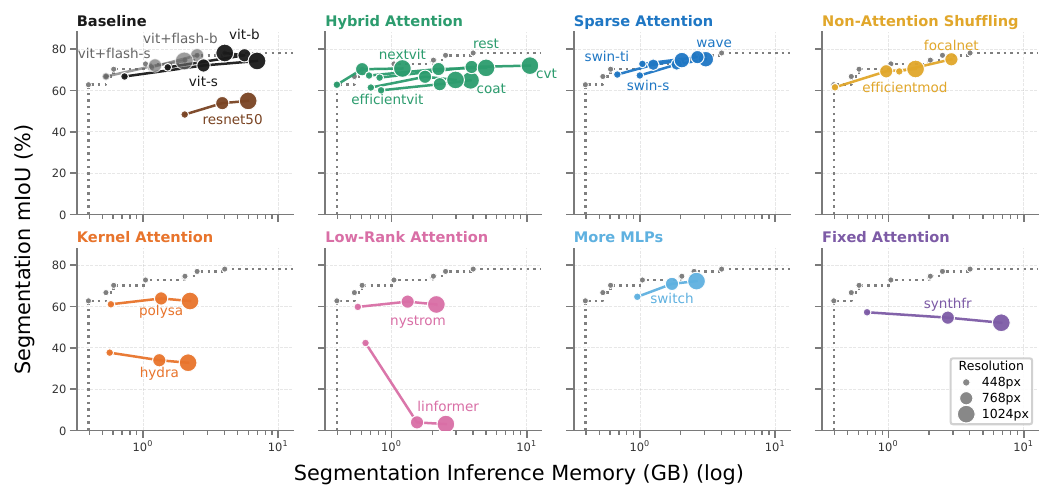}
		\caption{Inference memory vs.\ Cityscapes mIoU.}
	\end{subfigure}
	\caption{\textbf{Efficiency on Cityscapes.}
		Conventions as in \Cref{fig:eff-ade-sem}.
		At Cityscapes' $448$--$1024$\,px the collapse of the sequence-length-bound mixers is more severe than on ADE20K.
		NextViT and sparse models hold the memory Pareto-front.}
	\label{fig:eff-city-sem}
\end{figure}

\begin{figure}
	\begin{subfigure}{\textwidth}
		\centering
		\includegraphics[width=.8\textwidth]{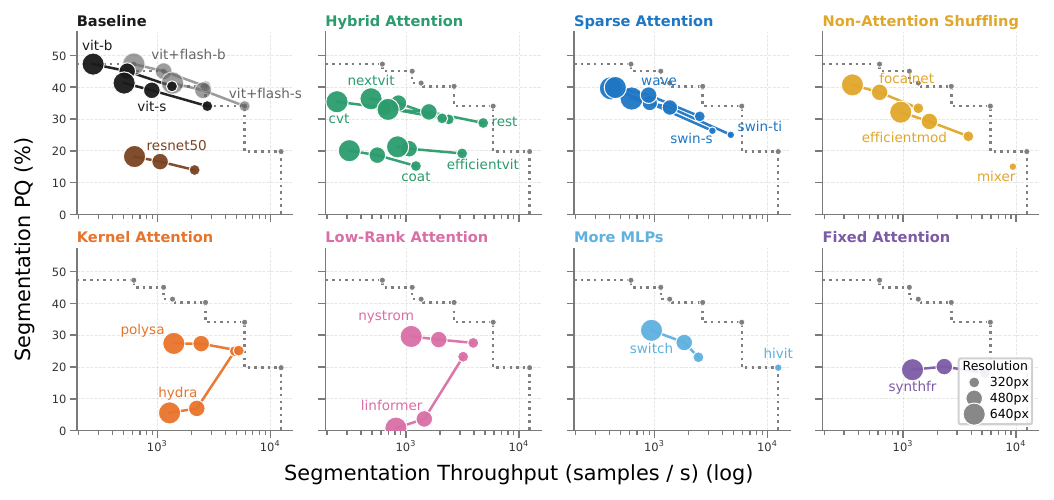}
		\caption{Throughput vs.\ COCO PQ.}
	\end{subfigure}
	\begin{subfigure}{\textwidth}
		\centering
		\includegraphics[width=.8\textwidth]{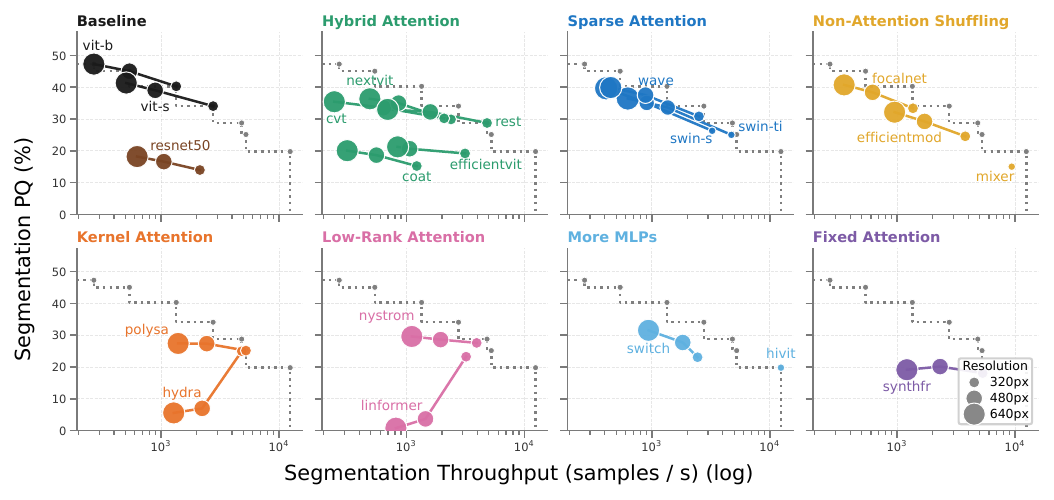}
		\caption{Throughput vs.\ COCO PQ \textbf{without FlashAttention}.}
	\end{subfigure}
	\begin{subfigure}{\textwidth}
		\centering
		\includegraphics[width=.8\textwidth]{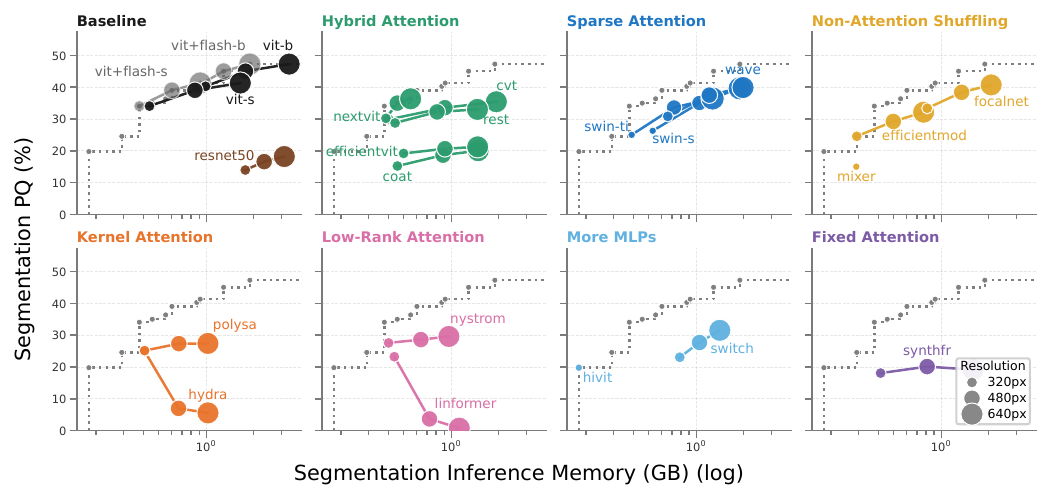}
		\caption{Inference memory vs.\ COCO PQ.}
	\end{subfigure}
	\caption{\textbf{Efficiency on COCO panoptic segmentation.}
		Conventions as in \Cref{fig:eff-ade-sem}.
		The ordering carries over from the semantic tasks. Without FlashAttention, ResT and HiViT are throughput-Pareto-optimal. NextViT is again memory-Pareto-optimal.}
	\label{fig:eff-coco-pano}
\end{figure}

\begin{figure}
	\begin{subfigure}{\textwidth}
		\centering
		\includegraphics[width=.8\textwidth]{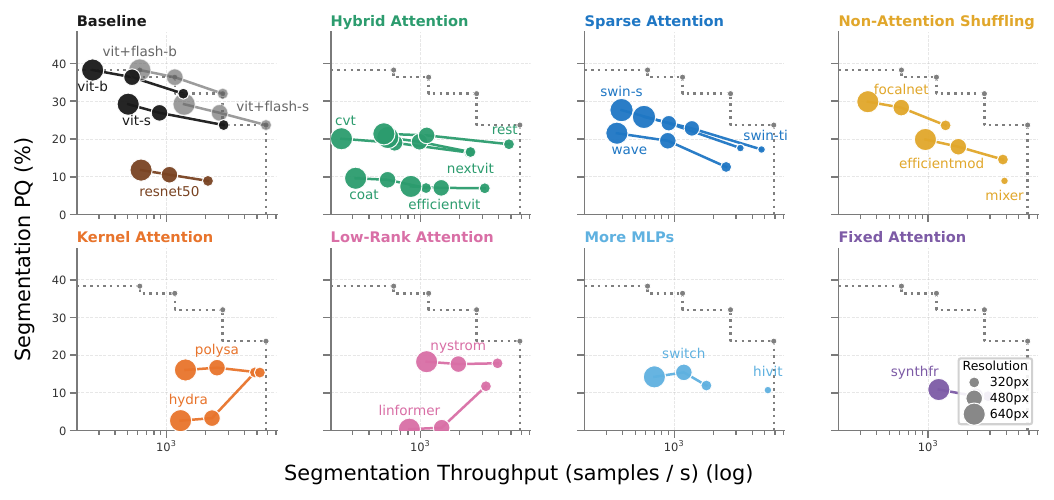}
		\caption{Throughput vs.\ ADE20K PQ.}
	\end{subfigure}
	\begin{subfigure}{\textwidth}
		\centering
		\includegraphics[width=.8\textwidth]{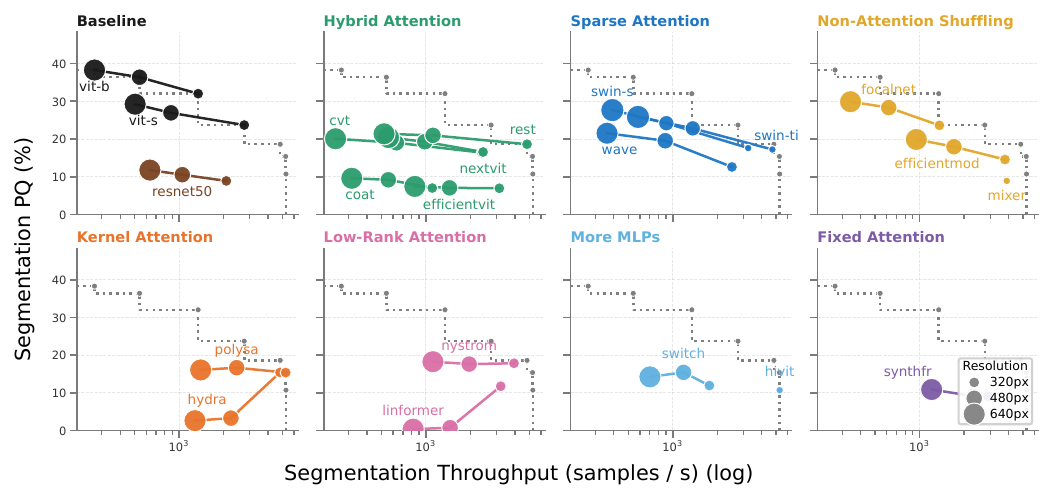}
		\caption{Throughput vs.\ ADE20K PQ \textbf{without FlashAttention}.}
	\end{subfigure}
	\begin{subfigure}{\textwidth}
		\centering
		\includegraphics[width=.8\textwidth]{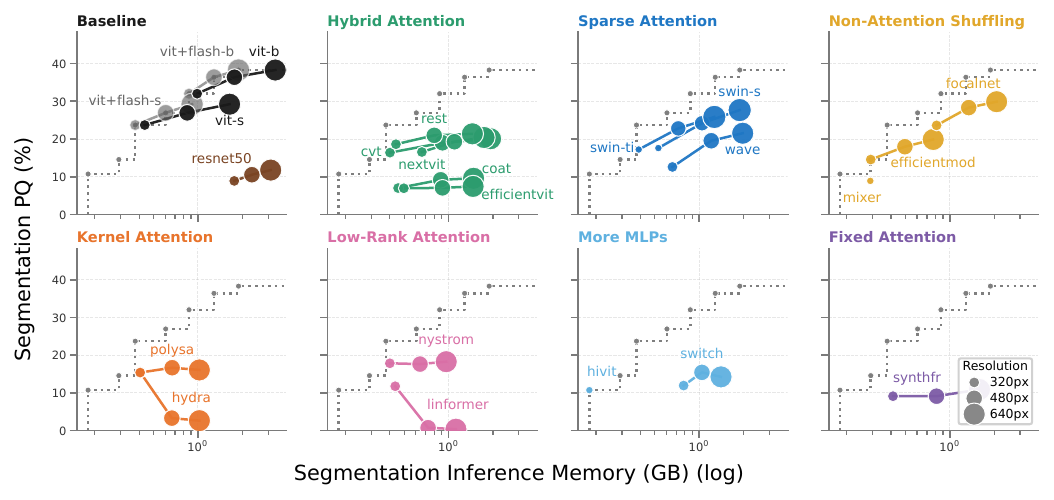}
		\caption{Inference memory vs.\ ADE20K PQ.}
	\end{subfigure}
	\caption{\textbf{Efficiency on ADE20K panoptic segmentation.}
		Conventions as in \Cref{fig:eff-ade-sem}. No efficient backbone reaches the segmentation quality of ViT.}
	\label{fig:eff-ade-pano}
\end{figure}

\Cref{fig:eff-ade-sem,fig:eff-city-sem,fig:eff-coco-pano,fig:eff-ade-pano} give both efficiency axes on all four tasks.
The same families hold the front on every task, and the same ones drop off it as resolution grows, so the verdict does not depend on the task or on which budget is charged.

\FloatBarrier
\subsection{Pre-training}
\label{apdx:more-plots-pretrain}

\begin{figure}
	\begin{subfigure}{.49\columnwidth}
		\centering
		\includegraphics[width=.95\textwidth]{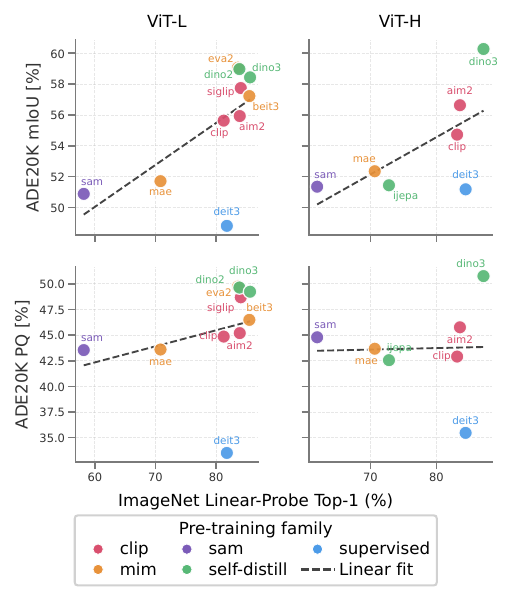}
		\caption{Pretraining quality on ADE20K semantic and panoptic.}
	\end{subfigure}
	\hfill
	\begin{subfigure}{.49\columnwidth}
		\centering
		\includegraphics[width=.95\textwidth]{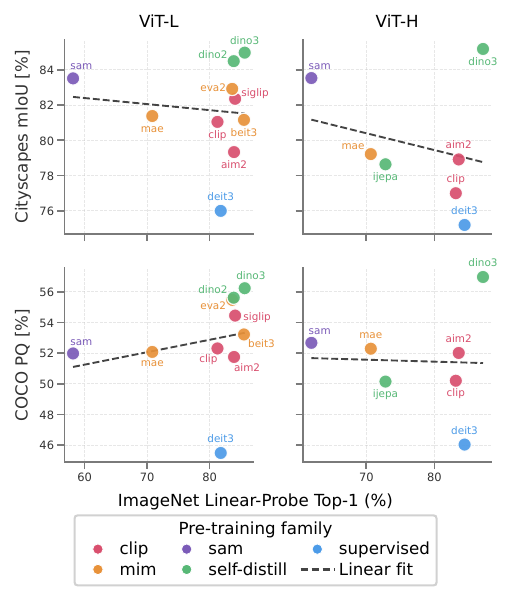}
		\caption{Pretraining quality on Cityscapes semantic and COCO panoptic.}
	\end{subfigure}
	\caption{\textbf{Pretraining objective against ImageNet linear-probe.}
		The Cityscapes fit is negative at both sizes; ADE20K panoptic is weakly positive for ViT-L and flat for ViT-H.
		Panoptic segmentation fits show no significant correlation between both tasks.
		Supervised DeiT~III sits far below the fit despite a competitive probe in every setting.}
	\label{fig:linprobe-grid-appendix}
\end{figure}

\FloatBarrier
\subsection{Compute Scaling}
\label{apdx:more-plots-scaling}

\begin{figure}
	\begin{subfigure}{.49\textwidth}
		\centering
		\includegraphics[width=.9\textwidth]{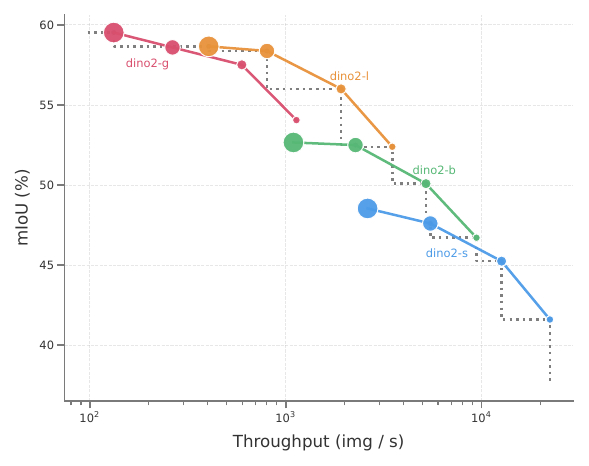}
		\caption{Throughput vs.\ ADE20K mIoU.}
	\end{subfigure}
	\hfill
	\begin{subfigure}{.49\textwidth}
		\centering
		\includegraphics[width=.9\textwidth]{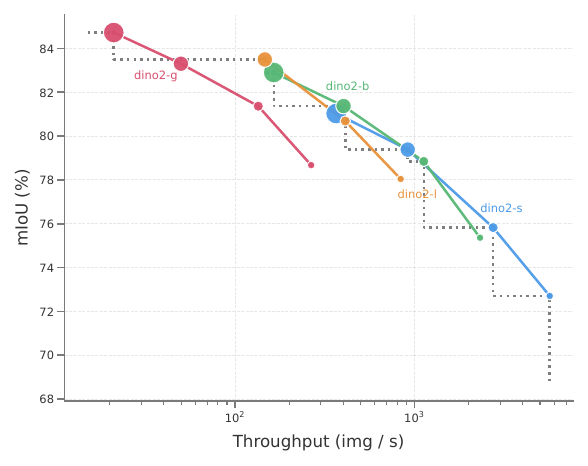}
		\caption{Throughput vs.\ Cityscapes mIoU.}
	\end{subfigure}
	\\
	\begin{subfigure}{.49\textwidth}
		\centering
		\includegraphics[width=.9\textwidth]{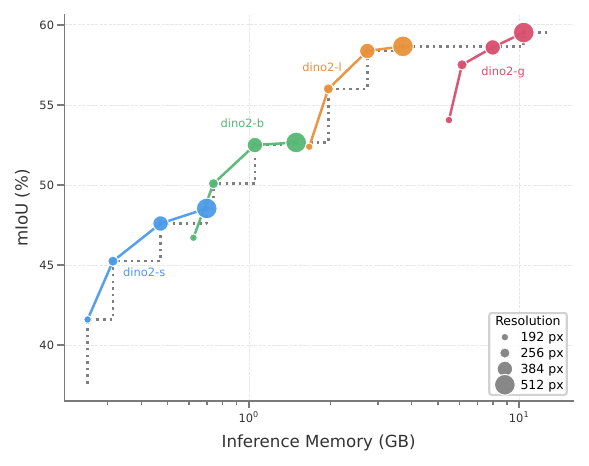}
		\caption{Inference memory vs.\ ADE20K mIoU.}
	\end{subfigure}
	\hfill
	\begin{subfigure}{.49\textwidth}
		\centering
		\includegraphics[width=.9\textwidth]{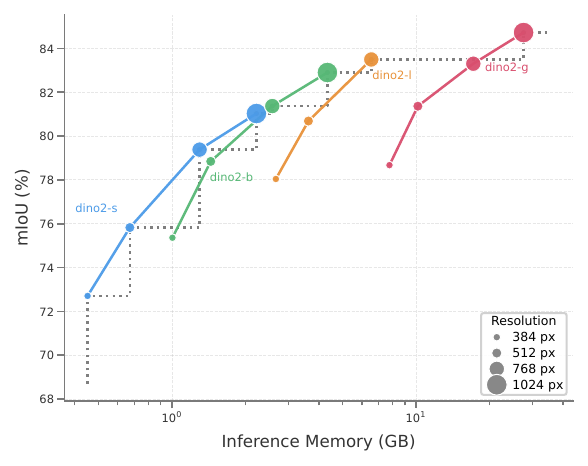}
		\caption{Inference memory vs.\ Cityscapes mIoU.}
	\end{subfigure}
	\caption{\textbf{Compute-optimal scaling on the semantic tasks.}
		DINOv2-S/B/L/g across resolutions, with marker size encoding resolution and the dotted staircase marking the Pareto front.
		On Cityscapes image resolution plays a bigger role than on all other tasks and model size does not give drastic improvements, especially when considering memory. This is in contrast to the three other tasks.}
	\label{fig:scaling-grid-sem}
\end{figure}

\begin{figure}
	\begin{subfigure}{.49\textwidth}
		\centering
		\includegraphics[width=.9\textwidth]{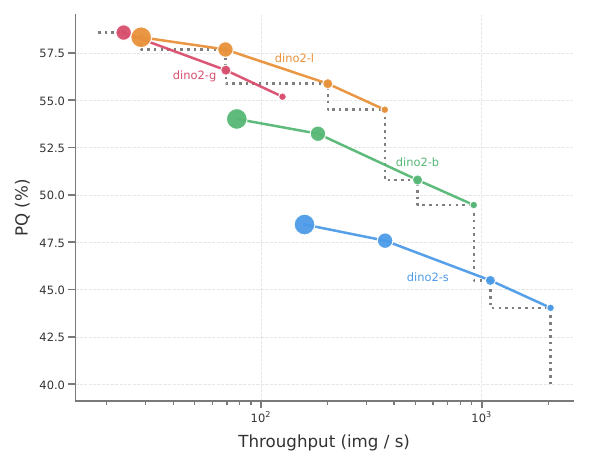}
		\caption{Throughput vs.\ COCO PQ.}
	\end{subfigure}
	\hfill
	\begin{subfigure}{.49\textwidth}
		\centering
		\includegraphics[width=.9\textwidth]{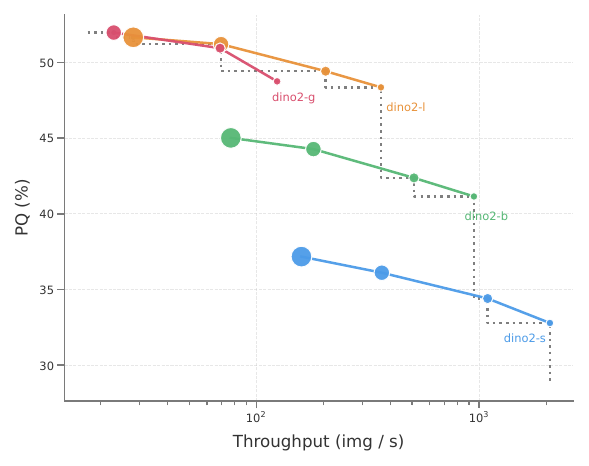}
		\caption{Throughput vs.\ ADE20K PQ.}
	\end{subfigure}
	\\
	\begin{subfigure}{.49\textwidth}
		\centering
		\includegraphics[width=.9\textwidth]{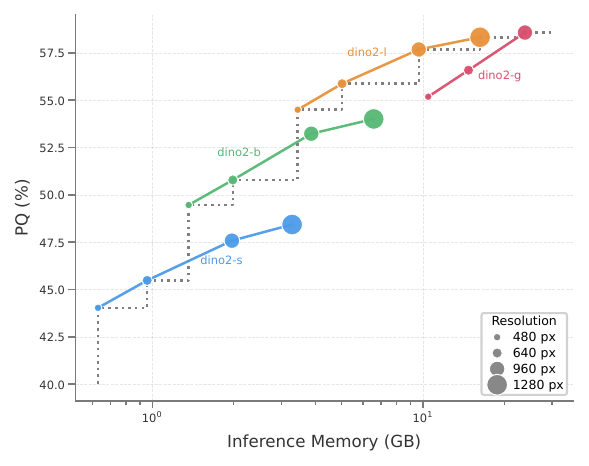}
		\caption{Inference memory vs.\ COCO PQ.}
	\end{subfigure}
	\hfill
	\begin{subfigure}{.49\textwidth}
		\centering
		\includegraphics[width=.9\textwidth]{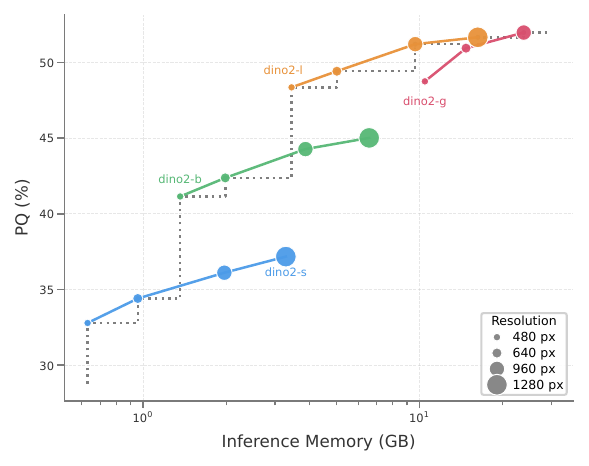}
		\caption{Inference memory vs.\ ADE20K PQ.}
	\end{subfigure}
	\caption{\textbf{Compute-optimal scaling on the panoptic tasks.}
		Conventions as in \Cref{fig:scaling-grid-sem}.
		The size-before-resolution ordering of ADE20K semantic clearly holds here too.}
	\label{fig:scaling-grid-pano}
\end{figure}

\Cref{fig:scaling-grid-sem,fig:scaling-grid-pano} add the inference-memory axis and the two panoptic tasks to the throughput plot of the main paper (\Cref{fig:model-res-scaling}).
Memory matters because the compute-optimal fit of \Cref{apdx:optimal-scaling} is a FLOPs argument, and a memory-bound reading could have favoured the low-resolution regime instead.
It does not, since the same DINOv2-S/B/L/g progression traces the front on both axes and all four tasks.

\end{document}